%% file: camera_ready.tex
\documentclass[twoside]{article}

\usepackage[accepted]{aistats2024}
\usepackage[round]{natbib}

\usepackage{algorithm}
\usepackage{algpseudocode}
\usepackage{amsmath}
\usepackage{amssymb}
\usepackage{amsthm}
\usepackage{booktabs}
\usepackage{graphicx}
\usepackage{subcaption}
\usepackage{tikz}
\usetikzlibrary{bayesnet}
\usepackage{url}

\newtheorem{proposition}{Proposition}

\makeatletter
\def\adl@drawiv#1#2#3{%
        \hskip.5\tabcolsep
        \xleaders#3{#2.5\@tempdimb #1{1}#2.5\@tempdimb}%
                #2\z@ plus1fil minus1fil\relax
        \hskip.5\tabcolsep}
\newcommand{\cdashlinelr}[1]{%
  \noalign{\vskip\aboverulesep
           \global\let\@dashdrawstore\adl@draw
           \global\let\adl@draw\adl@drawiv}
  \cdashline{#1}
  \noalign{\global\let\adl@draw\@dashdrawstore
           \vskip\belowrulesep}}
\makeatother

\begin{document}

\runningtitle{Towards Generalizable and Interpretable Motion Prediction: A Deep Variational Bayes Approach}


\twocolumn[

\aistatstitle{Towards Generalizable and Interpretable Motion Prediction: \\ A Deep Variational Bayes Approach}

\aistatsauthor{
  Juanwu Lu
  \And Wei Zhan
  \And Masayoshi Tomizuka
  \And Yeping Hu
}

\aistatsaddress{
  Purdue University \\ \texttt{juanwu@purdue.edu}
  \And UC Berkeley \\ \texttt{wzhan@berkeley.edu}
  \And UC Berkeley \\ \texttt{tomizuka@berkeley.edu}
  \And LLNL \\ \texttt{yeping\_hu@berkeley.edu}
}

]

% Abstract
\begin{abstract}
    \input{sections/abstract}   
\end{abstract}
% Introduction
\section{INTRODUCTION}
\input{sections/introduction}
% Related Work
\section{RELATED WORKS}
\input{sections/related}
% Methodology
\section{METHOD}

\input{sections/method}
% Experiment

\section{EXPERIMENTS}
\input{sections/experiment}
% Conclusion and Future Work
\section{CONCLUSION}
\input{sections/conclusion}

% If the paper has no references, uncomment the following line
% \nocite{*}
\bibliography{references.bib}

%%%%%%%%%%%%%%%%%%%%%%%%%%%%%%%%%%%%%%%%%%%%%%%%%%%%%%%%%%%%
\section*{Checklist}

% %%% BEGIN INSTRUCTIONS %%%
% The checklist follows the references. For each question, choose your answer from the three possible options: Yes, No, Not Applicable.  You are encouraged to include a justification to your answer, either by referencing the appropriate section of your paper or providing a brief inline description (1-2 sentences). 
% Please do not modify the questions.  Note that the Checklist section does not count towards the page limit. Not including the checklist in the first submission won't result in desk rejection, although in such case we will ask you to upload it during the author response period and include it in camera ready (if accepted).

% \textbf{In your paper, please delete this instructions block and only keep the Checklist section heading above along with the questions/answers below.}
% %%% END INSTRUCTIONS %%%

 \begin{enumerate}

 \item For all models and algorithms presented, check if you include:
 \begin{enumerate}
   \item A clear description of the mathematical setting, assumptions, algorithm, and/or model. \textbf{Yes}
   \item An analysis of the properties and complexity (time, space, sample size) of any algorithm. \textbf{Not Applicable}
   \item (Optional) Anonymized source code, with specification of all dependencies, including external libraries. \textbf{Yes}
 \end{enumerate}

 \item For any theoretical claim, check if you include:
 \begin{enumerate}
   \item Statements of the full set of assumptions of all theoretical results. \textbf{Yes}
   \item Complete proofs of all theoretical results. \textbf{Not Applicable}
   \item Clear explanations of any assumptions. \textbf{Yes}
 \end{enumerate}

 \item For all figures and tables that present empirical results, check if you include:
 \begin{enumerate}
   \item The code, data, and instructions needed to reproduce the main experimental results (either in the supplemental material or as a URL). \textbf{Yes}
   \item All the training details (e.g., data splits, hyperparameters, how they were chosen). \textbf{Yes}
         \item A clear definition of the specific measure or statistics and error bars (e.g., with respect to the random seed after running experiments multiple times). \textbf{Yes}
         \item A description of the computing infrastructure used. (e.g., type of GPUs, internal cluster, or cloud provider). \textbf{Yes}
 \end{enumerate}

 \item If you are using existing assets (e.g., code, data, models) or curating/releasing new assets, check if you include:
 \begin{enumerate}
   \item Citations of the creator If your work uses existing assets. \textbf{Yes}
   \item The license information of the assets, if applicable. \textbf{Not Applicable}
   \item New assets either in the supplemental material or as a URL, if applicable. \textbf{Yes}
   \item Information about consent from data providers/curators. \textbf{Not Applicable}
   \item Discussion of sensible content if applicable, e.g., personally identifiable information or offensive content. \textbf{Not Applicable}
 \end{enumerate}

 \item If you used crowdsourcing or conducted research with human subjects, check if you include:
 \begin{enumerate}
   \item The full text of instructions given to participants and screenshots. \textbf{Not Applicable}
   \item Descriptions of potential participant risks, with links to Institutional Review Board (IRB) approvals if applicable. \textbf{Not Applicable}
   \item The estimated hourly wage paid to participants and the total amount spent on participant compensation. \textbf{Not Applicable}
 \end{enumerate}

 \end{enumerate}

\onecolumn
\appendix
\section{Overview}
In this supplementary material, we provide detailed derivation about the training objective in section~\ref{sec: derivation}. Further, we cover more implementation details in section~\ref{sec: implementation-details} and showcase additional empirical results in section~\ref{sec: add-expr-res}.

\section{Derivations of Training Objectives}
\label{sec: derivation}
In this section, we provide detailed mathematics of the variational mixture of Gaussians with Normal-Wishart prior and the derivation of z-posterior and training objectives.
\subsection{Conjugate Prior}

In the GNeVA model, we leverage the Normal-Wishart distribution as the prior for the mean $\boldsymbol\mu$ and precision $\boldsymbol\Lambda$ the Gaussian distribution. The distribution is in the form
\begin{equation}
    p(\boldsymbol\mu,\boldsymbol\Lambda\mid\eta,\beta,V,\nu)=\mathcal{N}(\boldsymbol\mu\mid\eta,(\beta\Lambda)^{-1})\mathcal{W}(\boldsymbol\Lambda\mid V,\nu).
\end{equation}
Herein, $\mu$ and $V$ are the \textit{prior mean} and the positive-definite \textit{prior scale matrix}, while $\beta$ and $\nu$ control the strengths of belief in two priors, respectively. The density of the Normal-Wishart distribution is a product of a conditional normal distribution for the mean variable and a Wishart distribution for the precision
\begin{equation}
    \begin{aligned}
        \mathcal{N}(\mu\mid\eta,(\beta\Lambda)^{-1}) &= \frac{\beta^{\frac{D}{2}\det(\Lambda)}}{2\pi^{\frac{D}{2}}}\exp\left\{-\frac{\beta}{2}(\boldsymbol\mu-\eta)^\intercal\Lambda(\boldsymbol\mu-\eta)\right\}, \\
        \mathcal{W}(\Lambda\mid V,\nu)&=\frac{\det(\Lambda)^{\frac{\nu-D-1}{2}}\exp\left\{-\frac{1}{2}\text{trace}(\Lambda V^{-1})\right\}}{2^{\frac{\nu D}{2}}\Gamma_{D}\left(\frac{\nu}{2}\right)\det(V)^{\frac{\nu}{2}}},
    \end{aligned}
\end{equation}
where $\Gamma_D(\cdot)$ is the \textit{multivariate gamma function} and $D$ is the dimensionality of the normal variable.
\subsection{KL Divergence}

The general objective for training the GNeVA model is to maximize the probability of ground-truth goal $g$. Following the notations, we have the log-probability given by 
\begin{equation}
    \begin{aligned}
        \log p(g) &= \log\int_{\boldsymbol{\mu}}\int_{\boldsymbol{\Lambda}}\int_z p(g,\boldsymbol{\mu},\boldsymbol{\Lambda},z)d\boldsymbol{\mu}d\boldsymbol{\Lambda}dz \\
        & = \log\int_{\boldsymbol{\mu}}\int_{\boldsymbol{\Lambda}}\int_z p(g,\boldsymbol{\mu},\boldsymbol{\Lambda},z)\frac{q(z,\boldsymbol{\mu},\boldsymbol{\Lambda})}{q(z,\boldsymbol{\mu},\boldsymbol{\Lambda})}d\boldsymbol{\mu}d\boldsymbol{\Lambda}dz \\
        & = \log\mathbb{E}_{q(z,\boldsymbol{\mu},\boldsymbol{\Lambda}\mid g)}\left[\frac{p(g,\boldsymbol{\mu},\boldsymbol{\Lambda},z)}{q(z,\boldsymbol{\mu},\boldsymbol{\Lambda})}\right]
    \end{aligned}
\end{equation}

According to the Jensen's Inequality,
\begin{equation}
    \log\mathbb{E}_{q(z,\boldsymbol{\mu},\boldsymbol{\Lambda})}\left[\frac{p(g,\boldsymbol{\mu},\boldsymbol{\Lambda},z)}{q(z,\boldsymbol{\mu},\boldsymbol{\Lambda})}\right]\geq\mathbb{E}_{q(z,\boldsymbol{\mu},\boldsymbol{\Lambda})}\left[\log\frac{p(g,\boldsymbol{\mu},\boldsymbol{\Lambda},z)}{q(z,\boldsymbol{\mu},\boldsymbol{\Lambda})}\right].
\end{equation}

Hence, we can reduce the maximizing of log-probability to maximize its evidence lower bound (ELBO). Recall that our variational posterior distribution is given by $q(\boldsymbol{z,\mu,\Lambda})=q(\boldsymbol\mu\mid\boldsymbol\Lambda)q(\boldsymbol\Lambda)q(\boldsymbol{z\mid g,\mu,\Lambda})$, where we drop the observation variables for simplicity. Therefore, the ELBO is futher given by
\begin{equation}
    \label{eq: elbo}
    \begin{aligned}
        \mathcal{L}_\text{ELBO} &= \mathbb{E}_{q(z,\boldsymbol{\mu},\boldsymbol{\Lambda})}\left[\log\frac{p(g,\boldsymbol{\mu},\boldsymbol{\Lambda},z)}{q(z,\boldsymbol{\mu},\boldsymbol{\Lambda})}\right]\\
        & = \mathbb{E}_{q(z,\boldsymbol{\mu},\boldsymbol{\Lambda})}\log p(g\mid\boldsymbol{\mu},\boldsymbol{\Lambda},z)
        + \mathbb{E}_{q(\boldsymbol{\Lambda})}\int_{\boldsymbol{\mu}}\log\frac{p(\boldsymbol{\mu}\mid\boldsymbol{\Lambda})}{q(\boldsymbol{\mu}\mid\boldsymbol{\Lambda})}q(\boldsymbol{\mu}\mid\boldsymbol{\Lambda})d\boldsymbol{\mu} \\
        & + \int_{\boldsymbol{\Lambda}}\log\frac{p(\boldsymbol{\Lambda})}{q(\boldsymbol{\Lambda})}q(\boldsymbol{\Lambda})d\boldsymbol{\Lambda} + \mathbb{E}_{q(\boldsymbol{\mu},\boldsymbol{\Lambda})}\int_z\log\frac{p(z)}{q(z,\boldsymbol{\mu},\boldsymbol{\Lambda})}q(z,\boldsymbol{\mu},\boldsymbol{\Lambda})d\boldsymbol{\mu}d\boldsymbol{\Lambda} \\
        &= \mathbb{E}_{q(\boldsymbol{\mu},\boldsymbol{\Lambda})q(z)}\log p(g\mid\boldsymbol{\mu},\boldsymbol{\Lambda},z)
        -\mathbb{E}_{q(\boldsymbol{\Lambda})}D_{KL}\left(q(\boldsymbol{\mu}\mid\boldsymbol{\Lambda})\middle\|p(\boldsymbol{\mu}\mid\boldsymbol{\Lambda})\right) \\
        & - D_{KL}\left(q(\boldsymbol{\Lambda})\middle\|p(\boldsymbol{\Lambda})\right) - \mathbb{E}_{q(\boldsymbol{\mu},\boldsymbol{\Lambda})}D_{KL}\left(q(z\mid g,\boldsymbol\mu,\boldsymbol{\Lambda})\middle\|p(z)\right).
    \end{aligned}
\end{equation}

One benefit of using a conjugate prior is that it yields a closed-form solution to expectations in calculating the ELBO loss function. For the first term in equation~\ref{eq: elbo}, we can leverage the following property of multivariate normal distribution
\begin{equation}
    x\sim\mathcal{N}(\boldsymbol\mu,{\boldsymbol\Lambda}^{-1})\Rightarrow\mathbb{E}\left[x^\intercal\boldsymbol{A}x\right]=\text{trace}(\boldsymbol{A\Lambda^{-1}})+\boldsymbol{\mu^\intercal A\mu},
\end{equation}
which allows us to transform the first term in the form
\begin{equation}
    \label{eq: exp-emiss}
    \begin{aligned}
        \mathbb{E}_{q(z,\boldsymbol{\mu, \Lambda})}\log p(g\mid\boldsymbol{\mu,\Lambda},z)
        &=\frac{1}{2}\mathbb{E}_{q(z,\boldsymbol{\mu, \Lambda})}\left[-(g-\boldsymbol\mu)^\intercal\boldsymbol\Lambda(g-\boldsymbol\mu) - D\log(2\pi)+\log\det(\boldsymbol\Lambda)\right]z  \\
        &\propto -\frac{1}{2}\sum\limits_{c=1}^C q(z)\mathbb{E}_{q(\boldsymbol\Lambda_c)}\left[(g-\boldsymbol\mu_c)^\intercal\Lambda_c(g-\boldsymbol\mu_c)+\frac{D}{\beta_c}-\log\det(\Lambda)\right] \\
        &=-\frac{1}{2}\sum\limits_{c=1}^C q(z)\left[\nu_c(g-\boldsymbol\mu_c)^\intercal V_c(g-\boldsymbol\mu_c) + \frac{D}{\beta} -\log\det(V)-\psi_D(\frac{\nu_c}{2})+D\log\pi\right],
    \end{aligned}
\end{equation}
where $\psi_D(\cdot)$ is the \textit{multivariate digamma function}. Meanwhile, the second and the third term in the equation~\ref{eq: elbo} also have closed-form solutions~\citep{bishop2006pattern} in the form
\begin{equation}
    \mathbb{E}_{q(\boldsymbol{\Lambda})}D_{KL}\left(q(\boldsymbol{\mu}\mid\boldsymbol{\Lambda})\middle\|p(\boldsymbol{\mu}\mid\boldsymbol{\Lambda})\right) = \frac{1}{2}\sum\limits_{c=1}^C\beta_0\nu_c(\eta_c-\eta_0)^\intercal V_c(\eta_c-\eta_0) + \frac{K}{2}(\frac{\beta_0}{\beta_c}-\log\frac{\beta_0}{\beta_c}-1);
\end{equation}
\begin{equation}
    D_{KL}\left(q(\boldsymbol{\Lambda})\middle\|p(\boldsymbol{\Lambda})\right) = \frac{1}{2}\sum\limits_{c=1}^C\nu_c\left[\text{trace}(V_0^{-1}V_c)-D\right]-\nu_0\log\det(V_0^{-1}V_c)+2\log\frac{\Gamma_D(\frac{\nu_0}{2})}{\Gamma_D(\frac{\nu_c}{2})} + (\nu_c-\nu_0)\psi_D(\frac{\nu_c}{2}).
\end{equation}

\subsection{Variational $z$-posterior}

According to~\citet{bishop2006pattern}, the optimal factor of $\log q(z)$ is derived from
\begin{equation}
    \label{eq: opt-qz}
    \log q^*(z)=\mathbb{E}_{\boldsymbol{\mu,\Lambda}}\log p(g|\boldsymbol{\mu, \Lambda}, z) + \text{const}.
\end{equation}
Therefore, we can make use of the results in equation~\ref{eq: exp-emiss} and estimate the variational $z$-posterior by
\begin{equation}
    \begin{aligned}
        \log q(z=c) &=\log\pi_c + \frac{1}{2}\mathbb{E}_{q(\boldsymbol\Lambda_c)}[\log\det(\boldsymbol\Lambda_c)]-\frac{D}{2}\log(2\pi) - \frac{1}{2}\mathbb{E}_{q(\boldsymbol{\mu,\Lambda})}\left[(g-\boldsymbol\mu_c)^\intercal\boldsymbol\Lambda_c(g-\boldsymbol\mu_c)\right] \\
        &= \log\pi_c +\log\det(V_c) +\psi_2(\frac{\nu_c}{2}) - \frac{D}{2\beta_c} - \frac{\nu_c}{2}(g-\boldsymbol\mu_c)^\intercal V_c(g-\boldsymbol\mu_c) +\text{const}
    \end{aligned}
\end{equation}

\section{Implementation Details}
\label{sec: implementation-details}

\subsection{Model}
\label{sec: model-details}

\begin{figure}[!ht]
    \centering
    \begin{subfigure}{0.45\textwidth}
        \centering
        \caption{Context Attention Module}
        \includegraphics[width=\textwidth]{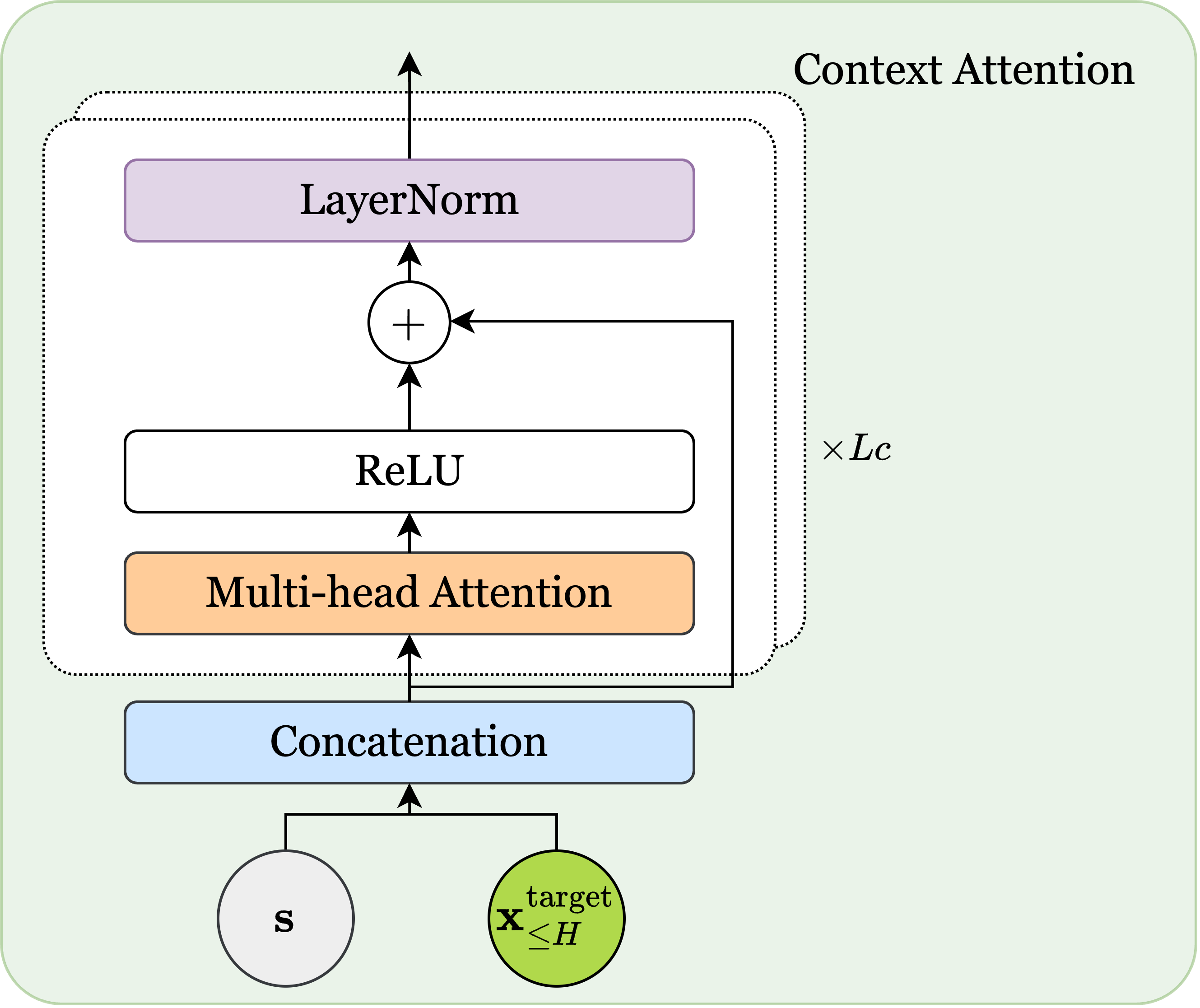}
        \label{fig: context-attention}   
    \end{subfigure}
    \hspace{0.1em}
    \begin{subfigure}{0.40\textwidth}
        \centering
        \caption{Interaction Attention Module}
        \includegraphics[width=\textwidth]{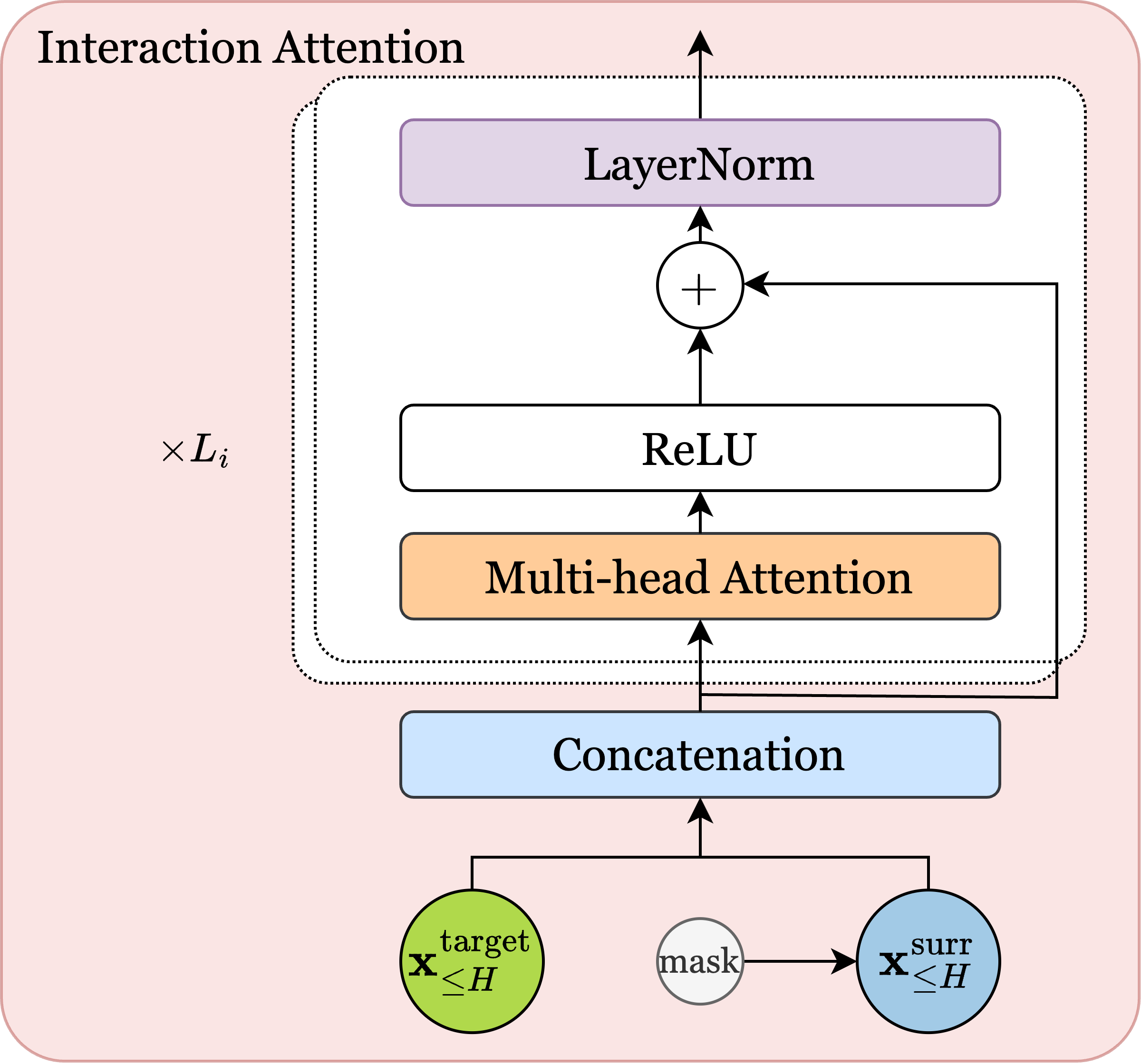}
        \label{fig: interaction-attention}   
    \end{subfigure}
    \caption{Illustrations of the Attention Modules.}
\end{figure}

We implement our model in PyTorch. All the Multi-layer Perceptrons (MLPs) inside the GNeVA model consist of a hidden layer and an output layer, with layer normalization and ReLU activation. The size of the hidden feature is set to 128. For parameterization of the goal spatial distribution, we have two attention-based modules that fuse features of the map and surrounding participants with the target vehicle, including a Context Attention Module and an Interaction Attention Module. The Context Attention module parameterizes the posterior distribution of means $q(\boldsymbol{\mu})$. As shown in Figure~\ref{fig: context-attention}, the module contains a cascade of $L_c$ self-attention layers.
% Suppose we have $N_\text{lane}$ lanes on the map and a total of $N_\text{agent}$ traffic participants.
The map feature map $\mathbf{s}$, target feature map $\mathbf{x}_{\leq H}^\text{target}$, and the surrounding participants feature map $\mathbf{x}_{\leq H}^\text{surr}$ concatenate to form the context feature map $\mathbf{X}_c$ and pass an multi-head attention (MHA) layer~\citep{NIPS2017_3f5ee243}. The operator is given as
\begin{equation}
    \begin{aligned}
        & \text{MHA}(\mathbf{X}) = \frac{\mathbf{Q}\mathbf{K}^\intercal}{\sqrt{d_k}}\mathbf{V}, \\
        \text{where}\quad & \mathbf{Q}=\mathbf{X}\mathbf{W}_q,\ \mathbf{K}=\mathbf{X}\mathbf{W}_k,\ \mathbf{V}=\mathbf{X}\mathbf{W}_v.
    \end{aligned}
\end{equation}
The $\mathbf{W}_q$, $\mathbf{W}_k$, and $\mathbf{W}_v$ are weights of linear layers. Following the MHA layer, there exists a ReLU activation, a residual connection layer, and a layer-normalization layer. The operation in a self-attention layer is given as
\begin{equation}
    \text{SelfAttention}(\mathbf{X}) = \text{LayerNorm}(\mathbf{X} + \text{ReLU}(\text{MHA}(\mathbf{X}))).
\end{equation}
In our implementation, we build our Context Attention module with $L_c=3$ self-attention layers. To get the target feature map after context attention, we extract the corresponding row in the output feature map that matches the locations of the target feature map in $\mathbf{X}_c$. It passes a consecutive MLP to derive the mean posterior parameters $\boldsymbol{\eta}^q$ and $\boldsymbol{\kappa}^q$.

As illustrated in Figure~\ref{fig: interaction-attention}, the Interaction Attention module shares a structure similar to the Context Attention module but differs in two ways. First, the input features are the feature maps of the target and surrounding participants. We drop the map features since they are temporally static; hence, we consider they have no effects on future uncertainty, as we mentioned in the paper. Meanwhile, we apply a mask to the surrounding participant feature map and remove those without potential future interactions with the target vehicle. In our implementation, we simplify the discussion by considering only agents with observations of their motion states at the observation horizon (i.e., $\mathbf{x}_H$) and remove the rest. We build our Interaction Attention module with $L_i=1$ self-attention layers and get the target feature map after interaction attention the same way as in the Context Attention module. 

\subsection{Training}
\label{sec: training-details}

We train our GNeVA model with a batch size of $64$ for $36$ epochs on two NVIDIA A100 GPUs. We use AdamW~\citep{loshchilov2019decoupled} with a weight decay of $0.001$. For scheduling, the learning rate increases linearly to $0.001$ for the first $1000$ steps, then cosine anneals to $3e-7$. In addition, we project raw coordinates in the source data into a target-centric reference frame by translating and then rotating the coordinates with respect to the location and heading angle of the target vehicle at the observation horizon.

% ---
% Additional experiment results
\section{Additional Experiment Results}
\label{sec: add-expr-res}

\subsection{Sensitivity Analysis: Road Geometry}
\label{sec: sens-road}

Road geometry plays a significant role in determining the spatial location of goal distributions. We expect the model to respond to changing road geometry to guarantee its generalization ability. In this experiment, we investigate how the proposed GNeVA model responds to the changing road geometry by adjusting the observation radius. A low observation radius leads to a reduced sensible range of surrounding lanes. We achieve this by removing all the lane polygons from the source data that have no intersection with the circle centered at the target vehicle's location at the observation horizon, with a radius of the observation radius. As a control factor, we assume complete observation of all the surrounding traffic under the same scenario.

\begin{figure}[!ht]
    \centering
    \begin{subfigure}{0.49\textwidth}
        \centering
        \caption{Map observation range radius $r=0$ meters}
        \includegraphics[width=\textwidth]{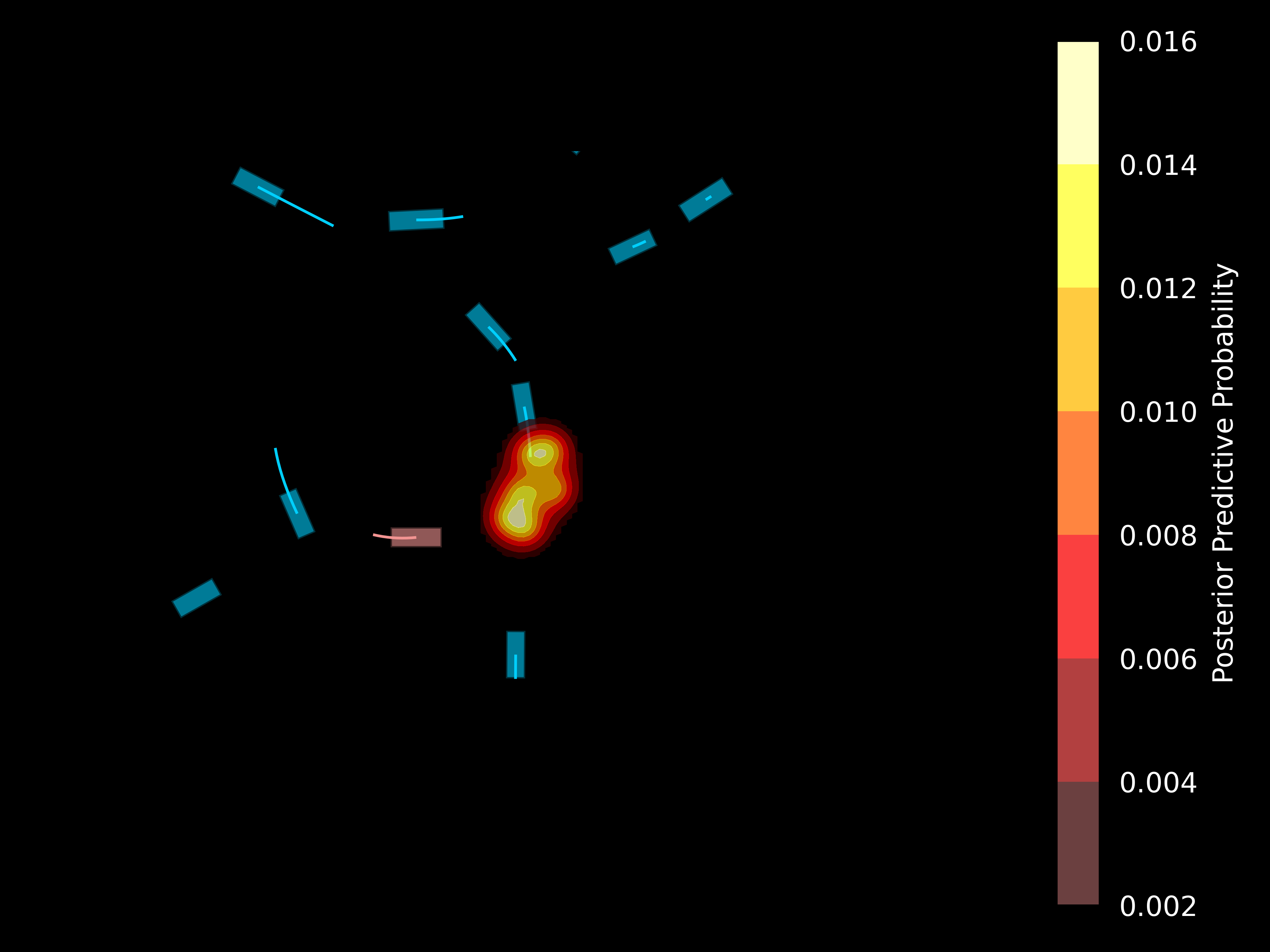}
        \label{fig: radius-0}
    \end{subfigure}
    \hspace{0.1em}
    \begin{subfigure}{0.49\textwidth}
        \centering
        \caption{Map observation range radius $r=5$ meters}
        \includegraphics[width=\textwidth]{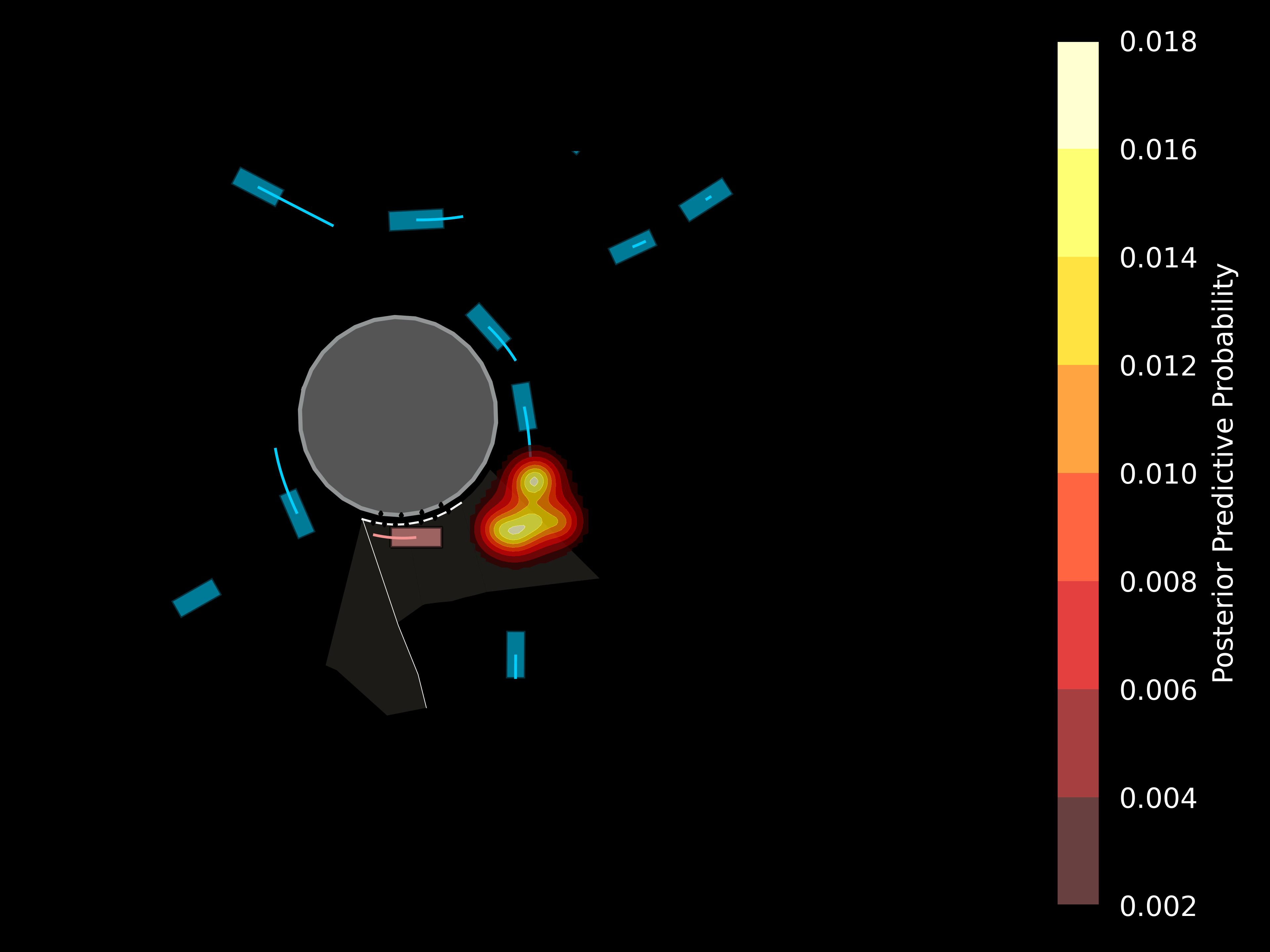}
        \label{fig: radius-5}
    \end{subfigure}
    \vfill
    \begin{subfigure}{0.49\textwidth}
        \centering
        \caption{Map observation range radius $r=10$ meters}
        \includegraphics[width=\textwidth]{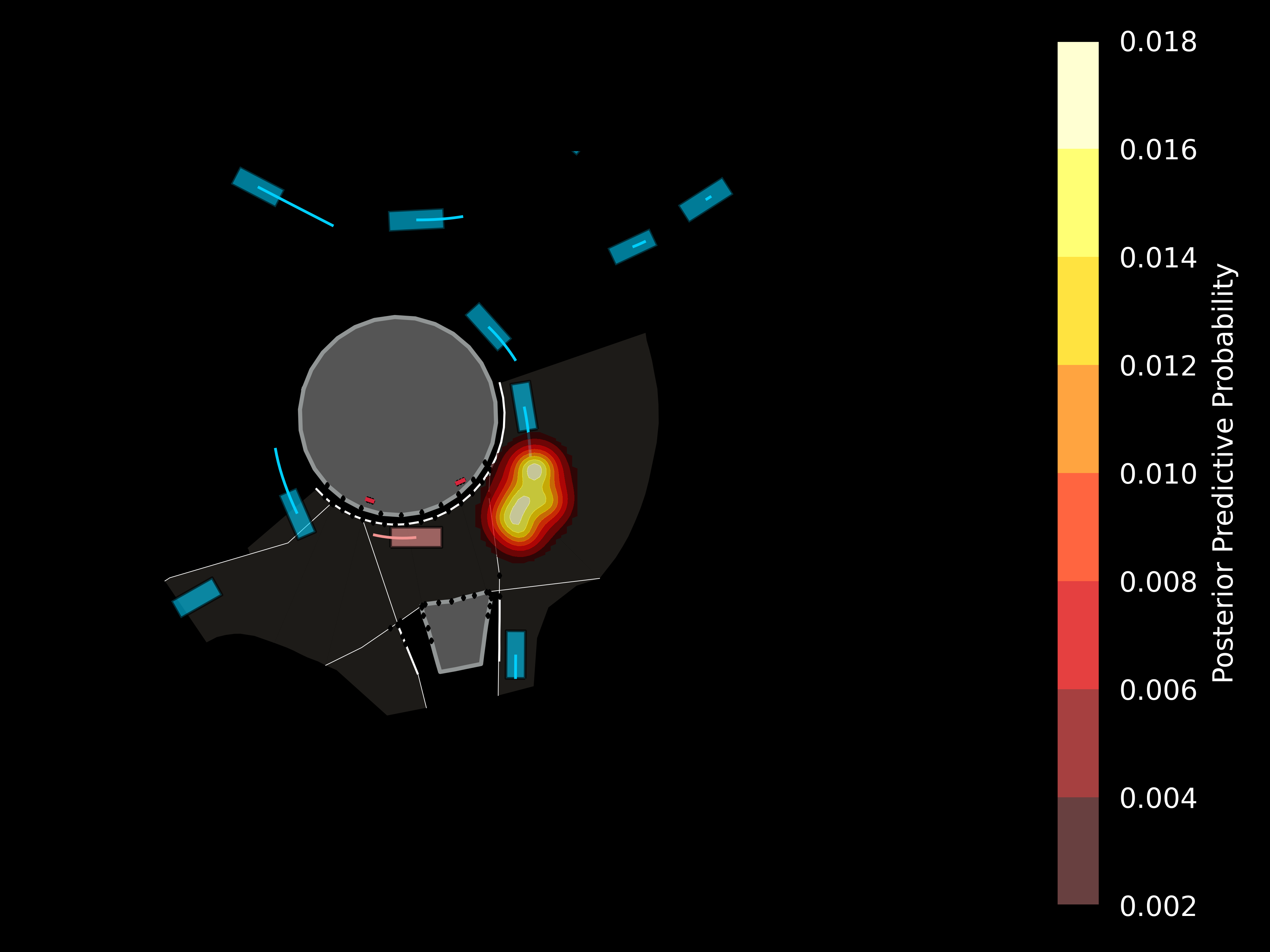}
        \label{fig: radius-10}
    \end{subfigure}
    \hspace{0.1em}
    \begin{subfigure}{0.49\textwidth}
        \centering
        \caption{Map observation range radius $r=20$ meters}
        \includegraphics[width=\textwidth]{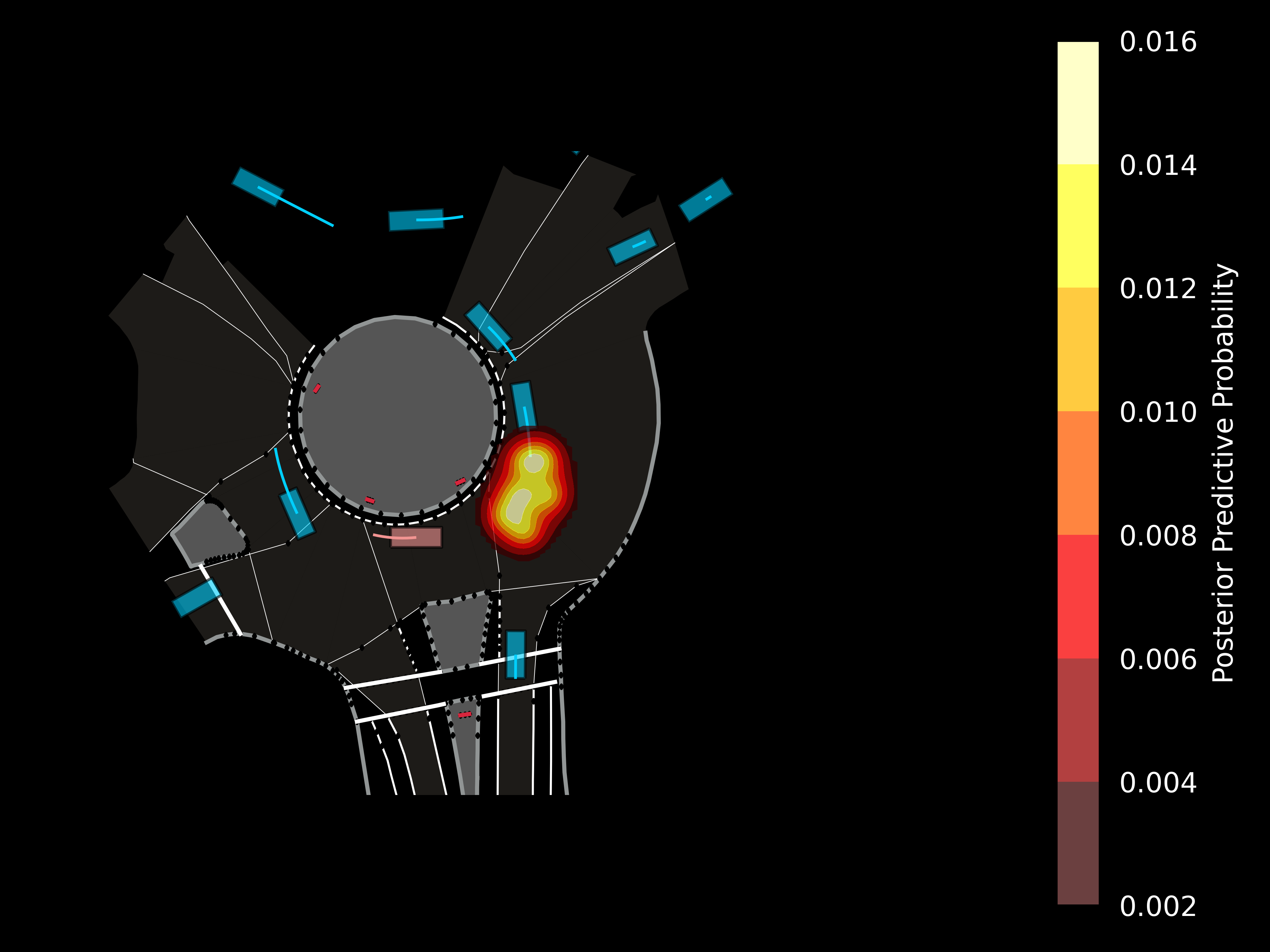}
        \label{fig: radius-20}
    \end{subfigure}
    \caption{Posterior predictive goal distribution under different map observation range radius settings. The case is sampled from the \texttt{DR\_USA\_Roundabout\_FT} scenario samples in the INTERACTION test dataset.}
    \label{fig: sensitivity}
\end{figure}

Results in Figure~\ref{fig: sensitivity} show that with increasing observation radius, the observation of the downstream lane structure becomes more accessible. Meanwhile, the shape of the posterior predictive goal distribution changes accordingly.

At first, when the radius is $r=0$ meters, the GNeVA model has no access to the map information, which means the model predicts the future movement of the target agent based solely on its trajectories and all surrounding participants. As a result, the problem resembles a car-following problem, where the agent follows the path of its leading vehicle. In the visualization, the predictive distribution is concentrated around the history trajectory of the nearest potential leading vehicle. The result indicates that the GNeVA model can learn the car-following maneuver unsupervised.

Then, when the radius gradually increases from $r=0$ to $r=10$ meters, the GNeVA model has more information about the shape of downstream lanes. We can observe the mixture components first spreading and then concentrating back around the history trajectory of the leading vehicle. This indicates that GNeVA can pay attention to the shape of the known lane. When $r=5$ meters, the model can only observe the current and the predecessor lane, where the current lane has a left boundary curved slightly towards the left and a right boundary slightly toward the right. Such a discrepancy potentially leads to the spreading of mixture components, indicating the target vehicle can go left or right. However, when $r=10$ meters, the model can fully observe the shape of the downstream lane, reducing uncertainty about the possible future.

Finally, when we increase the observation radius to $20$ and $50$ meters, the shape of the distribution does change quite significantly, which means the model has potentially learned only to pay attention to necessary downstream lanes. The visualization suggests that, instead of overfitting agents' history, the GNeVA model can effectively learn the road geometry constraints and assign proper attention to important lanes, which can be essential to guarantee its generalization ability~\citep{9981096}.

\subsection{Ablation Study: NMS Sampling}
\label{sec: ablation}

In this section, we investigate the effects of different NMS sampling settings on the final performance. Two key adjustable parameters of the NMS sampling algorithm are the \textbf{sampling radius} and the \textbf{sampling threshold}. The combined effect of these parameters determines the density of goal candidates sampled from the spatial distribution.

\paragraph{NMS Sampling Radius.} Table~\ref{tab: radius} shows the prediction performance evaluated under different sampling radius settings. Ranging from $0.5$ meters to $3.0$ meters with a fixed IoU threshold of 0\%, the optimal setting regarding mADE$_6$, mFDE$_6$, and MR$_6$ are observed at a radius of $2.5$ meters, $2$ meters, and $3$ meters, respectively. Generally, with an increasing sampling radius, the spread of goal candidates increases, and within a proper range, the probability of hitting the ground-truth goal increases.

\begin{table}[!ht]
\centering
\caption{Performance under different sampling radius. The results are evaluated on an IoU threshold of 0\%.}
\label{tab: radius}
\begin{tabular}{@{}cccc@{}}
\toprule
\textbf{Radius (meters)} & \textbf{mADE}   & \textbf{mFDE}   & \textbf{MR}       \\ \midrule
\textbf{0.5}    & 0.3238          & 0.7700            & 0.1058          \\
\textbf{1.0}    & 0.3111          & 0.7317          & 0.0953          \\
\textbf{1.5}    & 0.2994          & 0.6930          & 0.0850           \\
\textbf{2.0}    & 0.2952 & \textbf{0.6387} & 0.0780 \\
\textbf{2.5}    & \textbf{0.2946}          & 0.6730          & 0.0716          \\
\textbf{3.0}    & 0.2970          & 0.6764          & \textbf{0.0664}          \\ \bottomrule
\end{tabular}
\end{table}

\paragraph{NMS Sampling IoU Threshold.} Results in Table~\ref{tab: iou} indicate that with a fixed sampling radius, the prediction performance varies with different IoU thresholds. In our experiment, we find that with a sampling radius of $2$ meters, the optimal setting regarding mADE$_6$, mFDE$_6$, and MR$_6$ are observed at the threshold of $50\%$, $0\%$, and $25\%$. Generally, with an increasing IoU threshold, the spread of goal candidates decreases, leading to a denser candidate pool.

\begin{table}[!ht]
\centering
\caption{Performance under different IoU thresholds. The results are evaluated on a sampling radius of $2$ meters.}
\label{tab: iou}
\begin{tabular}{@{}cccc@{}}
\toprule
\textbf{IoU (\%)} & \textbf{mADE} & \textbf{mFDE} & \textbf{MR} \\ \midrule
\textbf{0}         & 0.2952        & \textbf{0.6387}        & 0.0780    \\
\textbf{25}         & 0.2965        & 0.6707        & \textbf{0.0614}    \\
\textbf{50}         & \textbf{0.2949}        & 0.6706        & 0.0651    \\ \bottomrule
\end{tabular}
\end{table}

In our paper, we balance the three metrics and use a radius of $2$ meters with $0\%$ IoU threshold.

\end{document}

%% file: sections/abstract.tex
Estimating the potential behavior of the surrounding human-driven vehicles is crucial for the safety of autonomous vehicles in a mixed traffic flow. Recent state-of-the-art achieved accurate prediction using deep neural networks. However, these end-to-end models are usually black boxes with weak interpretability and generalizability. This paper proposes the Goal-based Neural Variational Agent (\textit{GNeVA}), an interpretable generative model for motion prediction with robust generalizability to out-of-distribution cases. For interpretability, the model achieves target-driven motion prediction by estimating the spatial distribution of long-term destinations with a variational mixture of Gaussians. We identify a causal structure among maps and agents' histories and derive a variational posterior to enhance generalizability. Experiments on motion prediction datasets validate that the fitted model can be interpretable and generalizable and can achieve comparable performance to state-of-the-art results. 

%% file: sections/introduction.tex
\begin{figure}[!ht]
    \centering
    \includegraphics[width=\columnwidth]{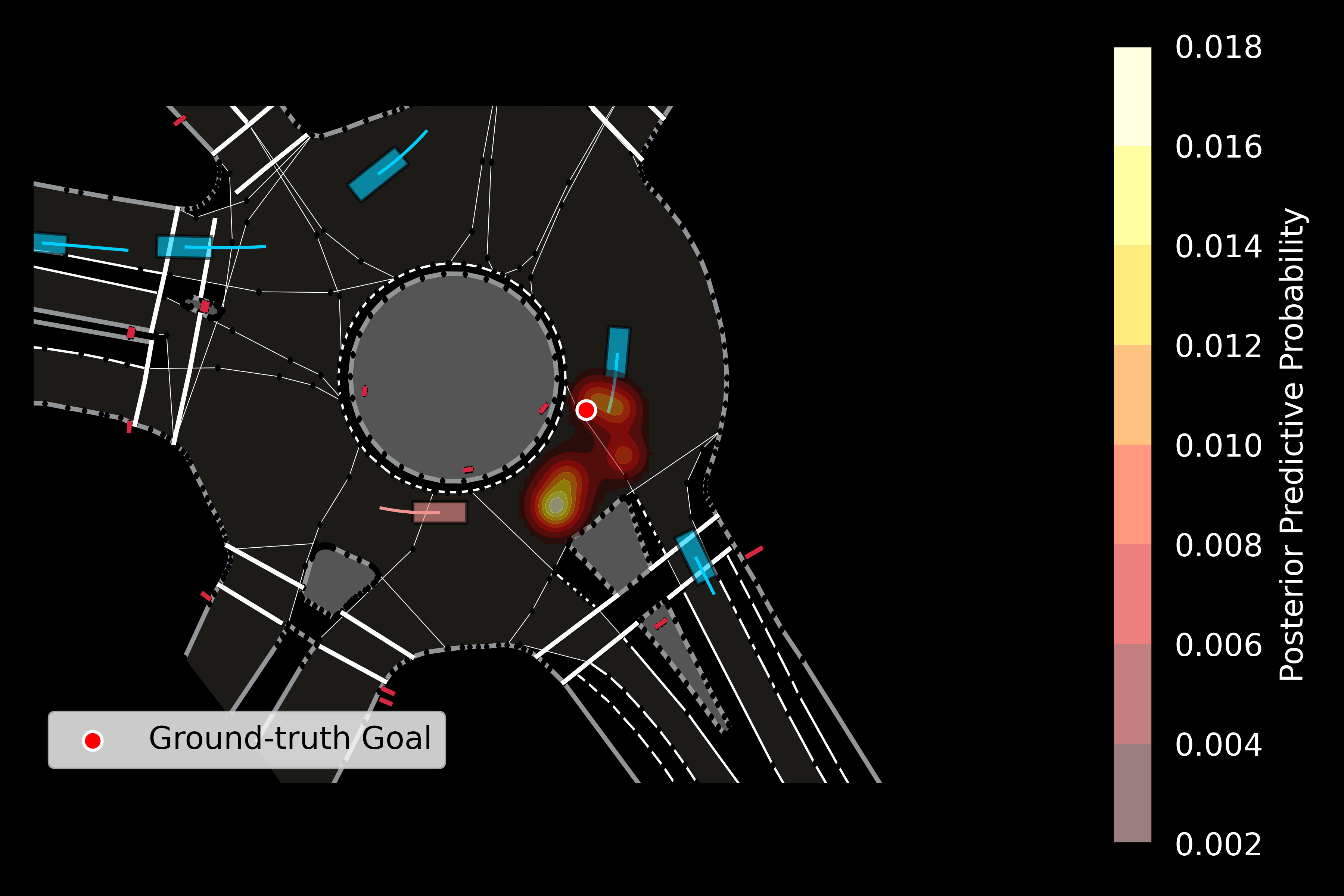}
    \caption{Example case illustrating the multi-modal distribution of long-term goal. The target vehicle can maintain its current cruising speed or accelerate inside the roundabout, leading to a spatial distribution with multiple modes.}
    \label{fig:multi_modal}
\end{figure}
With the rapid commercialization of autonomous vehicles (AVs), one can foresee increasing AV penetration rates and interactions among AVs and their surrounding human-driven vehicles (HDVs). To safely navigate through mixed and congested traffic, AVs must evaluate and predict future collision risks. A related task is motion prediction, which refers to forecasting the future trajectories of surrounding objects. However, motion prediction can be challenging due to uncertain and multi-modal driver behaviors: the exact history trajectories can point to several possible future paths, conditioned on plausible destinations, road geometry, social interactions, and maneuver differences, as illustrated in Figure~\ref{fig:multi_modal}.

Recently, learning-based methods have achieved outstanding prediction accuracy and gained emerging popularity. Nevertheless, most of these models assume that training and test data follow similar statistics and fit their parameters as point estimations derived from maximum likelihood. As a result, they can make over-confident predictions under distributional shifts due to changing road geometry or traffic conditions~\citep{balaji2017simple, filos2020can, Bahari_2022_CVPR}. Although meticulously adjusting the architecture, constantly enlarging the number of parameters, or training the model with a sufficiently large dataset that covers a diversified situation can help mitigate the performance degradation facing out-of-distribution (OOD) data~\citep{pmlr-v162-wang22u}, the time and computing resources required for data collection and parameter tuning can be intractable. Meanwhile, interpretability is necessary for the robustness and safety of deploying motion prediction algorithms in real-world cases. While end-to-end prediction models can deliver high accuracy, they are primarily black boxes with extremely limited interpretability.

In this work, we address these limitations and propose the Goal-based Neural Variational Agent (\textit{GNeVA}) model. The model follows a target-driven trajectory prediction setting~\citep{pmlr-v155-zhao21b, Gu_2021_ICCV}, where the motion prediction consists of two subtasks: predicting a continuous spatial distribution over plausible trajectory endpoints (i.e., \textit{goals}) on the map and completing the intermediate trajectory from the current location to the goals. The source code of our model is open-sourced at \url{https://github.com/juanwulu/gneva}. The main contribution of this paper is summarized as follows:

\begin{itemize}
    \item We identify and implement a causal structure where the surrounding physical context features determine the expected locations of goals, and the uncertainty is only sourced from dynamic future interactions.
    \item We propose a generative model using a variational mixture of Gaussians with learnable prior and variational posterior to model the spatial distribution of goals. The variational posterior is derived from the causal structure.
    \item We comprehensively evaluate GNeVA on the Argoverse Motion Forecasting~\citep{Argoverse} and the INTERACTION dataset~\citep {https://doi.org/10.48550/arxiv.1910.03088}, where the model is shown to yield comparable performance to the state-of-the-art motion prediction models. Crucially, the model maintains its performance under cross-scenario and cross-dataset tests, which indicates a promising generalizability. We further showcase predicted intention distributions qualitatively.
\end{itemize}

%% file: sections/related.tex
\begin{figure*}[!th]
    \centering
    \includegraphics[width=\textwidth]{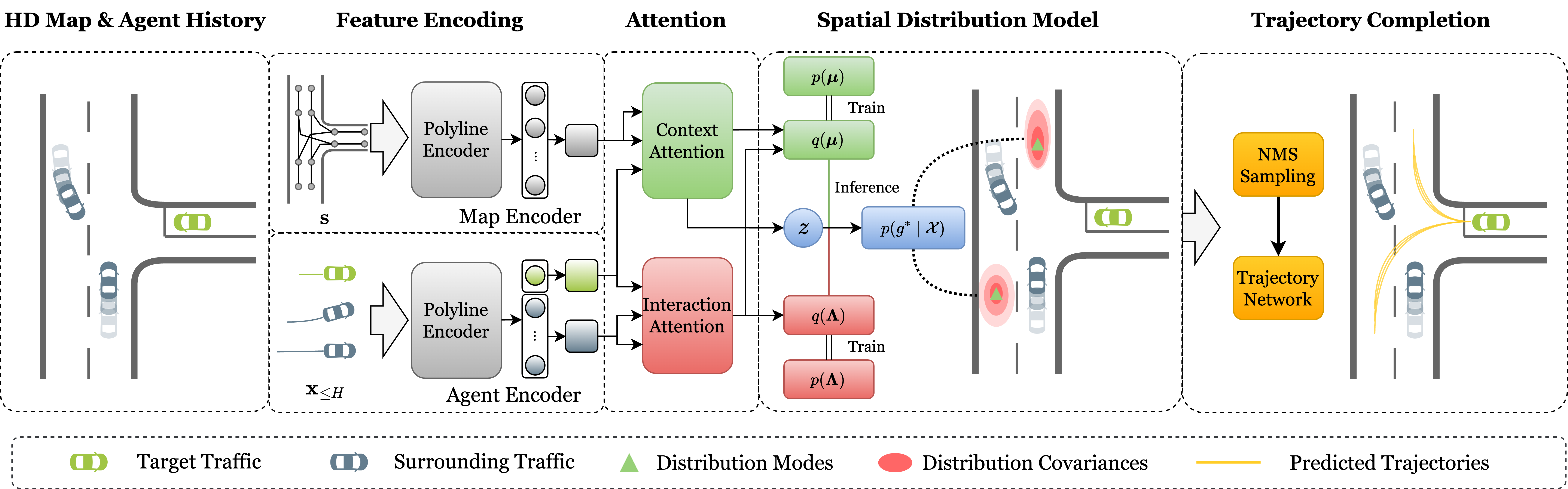}
    \caption{GNeVA model overview. The input HD map and history trajectories of all observed traffic are first encoded through polyline-based Map and Agent encoders, respectively. Encoded vector features associated with road geometry and agents' histories pass the context attention and interaction attention modules to derive posterior distribution parameters of means and precision. We then evaluate and sample goals from the posterior predictive distribution. Finally, a trajectory network completes the intermediate paths from current positions to sampled goals.}
    \label{fig: framework}
\end{figure*}
\subsection{Generalizability and Interpretability in Motion Prediction Models}
\label{sec:gen-int-mp}
While a few existing works implicitly consider generalization ability and interpretability in their model design~\citep{Tran_2021_WACV, Gu_2021_ICCV, 9636035, 9812253}, they did not thoroughly investigate these properties or clearly state what kind of design or model structure helped to address them. Other works tried to address these issues explicitly in motion prediction tasks. For example,~\citet{9901474} and~\citet{https://doi.org/10.48550/arxiv.2202.05140} proposed using domain-invariant semantic representation to minimize the joint discrepancy across traffic domains in a shared latent feature space. \citet{hu2022causal} utilized a causal structure to learn temporal invariant representations. These methods, however, still have limited interpretability on the latent correlations between output predictions and input features.

One way to incorporate interpretability is to depict the distribution of plausible future trajectories using a generative model. The classic approach employs flexible implicit distributions of trajectories from which the predictions can be drawn, such as conditional variational autoencoder~\citep{Lee_2017_CVPR} and generative adversarial networks~\citep{pmlr-v100-chai20a}. Despite their competitive prediction accuracy, using latent variables for reasoning driver behavior prohibits them from being interpreted and often requires inference-time sampling or ensembling methods to evaluate the uncertainty. Previous works address the limitation by decomposing trajectory prediction into two sub-tasks, intent identification and trajectory completion, based on the assumption that the uncertainties about a sufficiently long future trajectory can be mostly captured by its destination~\citep{Dendorfer_2021_ICCV}.

\subsection{Deep Generative Model}
\label{sec:deep-var-gen-model}
Driven by the emerging progress in generative models ~\citep{sohn2015learning, creswell2018generative}, there has been a significant shift in the paradigm of predictive modeling. Unlike deterministic regressors, generative models produce a distribution representing potential future behaviors. Notably, deep generative models, which exploit the expressiveness of neural networks for approximating the actual underlying probability distributions, have emerged as leading-edge methods. One primary deep generative approaches that stand out in the field are (Conditional) Variational Autoencoders ((C)VAEs)~\citep{kingma2013auto, sohn2015learning}, which applies variational inference through parameterizing the generative and variational family distributions with neural networks. Such methodologies have seen extensive application in predicting future vehicle trajectories within interactive scenarios~\citep{gupta2018social, lee2017desire, kosaraju2019social}. However, existing (C)VAE-based models commonly draw random samples from the latent space during inference, which has no guarantee for generalization, and the latent space lacks direct interpretability. Meanwhile, they often adopt uni-modal Gaussian distributions as both generative and variational distributions, which can fail to capture the multi-modality in future trajectories.

%% file: sections/method.tex
This section first presents the problem statements for the motion prediction problem (Section~\ref{sec: problem-statement}). Then, we introduce the formulation of the proposed deep generative model (Section~\ref{sec: spatial-model}) with causal structure. In this model, the prior distribution consists of trainable parameters, and the posterior distribution is parameterized by neural network encoders (Section~\ref{sec: feature-encoder}). To allow sampling from the multi-modal mixture model, we train a proxy network evaluating the mixture assignments (Section~\ref{sec: z-proxy}). Finally, we introduce how to sample from the posterior predictive distribution, generate intermediate trajectory(Section~\ref{sec:  sampling}), and train the model (Section~\ref{sec:  training}).

\input{sections/method/problem_statement}
\input{sections/method/spatial_distribution}
\input{sections/method/feature_encoder}
\input{sections/method/z_proxy}

\begin{algorithm}[!th]
\caption{Goal Sampling}
\label{alg:nms}
\begin{algorithmic}[1]
\Require List of candidate locations $G$ with associated probabilities.
\Require Candidate buffer radius $r$
\Require Intersection-over-Union (IoU) Threshold $\gamma$
\Ensure List of selected candidates $D$
\State Sort candidates by probabilities $p(g),g\in G$ in descending order
\State Initialize an empty list $D$
\While{$G \neq \emptyset$}
    \State Take the most probable candidate $g^*$ from $G$
    \State Add $g$ to $D$
    \State Create circle $c$ centered at $g$ of radius $r$
    \For{each candidate $g'$ in $G$}
        \State Create circle $c'$ centered at $g'$ of radius $r$
        \If{IoU$(c, c') > \gamma$}
            \State Remove $g'$ from $G$
        \EndIf
    \EndFor
    \State Remove $g$ from $G$
\EndWhile
\State \textbf{return} $D$
\end{algorithmic}
\end{algorithm}

\input{sections/method/sampling}
\input{sections/method/training}

%% file: sections/method/problem_statement.tex
\subsection{Problem Statement}
\label{sec: problem-statement}
In this paper, we follow the target-driven trajectory prediction problem setting and formulate the motion prediction as a two-step regression problem. The following lists out key concepts and the notations we use:
\begin{itemize}
   \item \textbf{Environment Semantics} refers to the objects in the surroundings besides traffic participants. Typical environment semantic data include high-resolution maps (HD maps), point clouds, and traffic regulations (e.g., stop signs, traffic lights, etc.). The set of all objects' indexes is denoted as $\mathcal{S}$, and the features of the $i$-th object observed at time step $t$ is denoted by $\mathbf{s}_t^{(i)},\ i\in\mathcal{S}$.
   \item \textbf{Traffic Participants} consists of individuals or entities interacting in the current traffic, such as pedestrians, cyclists, and pedestrians. The set of all traffic participants' indexes is denoted as $\mathcal{P}$, and the motion states of participant $i$ observed at time step $t$ is marked as $\mathbf{x}_t^{(i)},\ i\in\mathcal{P}$.
   \item \textbf{Target Participants} refers to a set of participants whose possible future locations are to be predicted. The collection of all target participants' indexes is denoted as $\mathcal{T},\ \mathcal{T}\subseteq\mathcal{P}$, and the motion states of participant $i$ observed at time step $t$ is similarly denoted as $\textbf{x}_t^{(i)},\ i\in\mathcal{T}$.
   \item \textbf{Observation Horizon} is the number of history time steps we have observed for prediction, denoted by $H$.
   \item \textbf{Prediction Horizon} is the number of future time steps to predict, denoted by $T$.
\end{itemize}

Suppose we aim to predict the future trajectory of $\left|\mathcal{T}\right|=N$ target vehicles. The objective is to search for an optimal model in model space $\mathcal{F}$ such that it maximizes the joint likelihood of future trajectories conditioned on the observations given by,
\begin{equation}
    \max_{f\in\mathcal{F}}\prod\limits_{i=1}^N\prod\limits_{t=1}^{T}p\left(\textbf{x}_{H+t}^{(i)}\mid f(\mathbf{x}_{<H+t}^{(j)}, \mathbf{s}_{<H+t}^{(k)})\right),
\end{equation}
where $i\in\mathcal{T}$, $j\in\mathcal{P}$, and $k\in\mathcal{S}$. However, direct estimation of the joint likelihood is challenging, and implicit estimations can limit interpretability. The target-driven trajectory prediction assumes that goals satisfy physical constraints and social interactions and capture the most uncertainty in a long-horizon prediction. We can reduce the estimation to maximize the upper bound of the original objective
\begin{equation}
    \max_{f^\prime\in\mathcal{F}}\prod\limits_{i=1}^Np\left(\textbf{x}_{H+T}^{(i)}\mid f(\mathbf{x}_{\leq H}^{(j)}, \mathbf{s}_{\leq H}^{(k)})\right).
\end{equation}

We further consider a more straightforward case of the problem: to predict the trajectory of a single agent (\textit{i.e.}, $N=1$) conditioned on its surrounding traffic participants and a static environment (\textit{i.e.}, $\mathbf{s}_t^{(i)}=\mathbf{s}^{(i)},\forall i\in\mathcal{S},t=1, \ldots, H+T$). To achieve this, we propose the GNeVA model (illustrated in Figure~\ref{fig: framework}) to learn and output the distribution representing the likelihood of an agent's goal at the prediction horizon. A trajectory network completes the intermediate path from its current location to the final goal. A spatial distribution model built with a variational mixture of Gaussians is the key component that drives the likelihood evaluation.

%% file: sections/method/spatial_distribution.tex
\subsection{Spatial Distribution Model for Goals}
\label{sec: spatial-model}
For simplicity, let $g\subset\mathbf{x}_{H+T}^{(i)}\in\mathbb{R}^2$ denote the two-dimensional location of the goal at the prediction horizon for a target participant $i$. Denote the collection of observed goals in the dataset by $\boldsymbol{g}=\{g_1,\ldots,g_N\}$. For each observation $g_n$, we can introduce a latent variable $z_n\in\{0, 1\}^C$. As illustrated in Figure~\ref{fig:likelihood-family}, we consider the generating process of observed sample $g_n$ in the form
\begin{equation}
    z_n\sim\text{Categorical}(\pi), 
\end{equation}
\begin{equation}
    g_n\mid z_n,\boldsymbol{\mu},\boldsymbol{\Lambda}\sim\prod\limits_{c=1}^C\mathcal{N}\left(g_n\mid\boldsymbol{\mu}_c,\boldsymbol{\Lambda}_c^{-1}\right)^{z_c},
\end{equation}
\begin{equation}
    \boldsymbol{\mu}_c,\boldsymbol\Lambda_c\sim\mathcal{N}\left(\boldsymbol\mu_c\mid\eta_0,\left(\beta_0\boldsymbol\Lambda_c\right)^{-1}\right)\mathcal{W}\left(\boldsymbol\Lambda_c\mid V_0,\nu_0\right).
\end{equation}
where $C$ is a tunable number of mixture components and $\pi$ is the mixing coefficients. The formulation addresses two propositions related to the nature of multi-modality and uncertainty of the goals:

\begin{figure}[!ht]
    \centering
    \begin{subfigure}{.4\columnwidth}
        \centering
        \begin{tikzpicture}
            % Nodes
            \node[obs] (g) {$g$}; %
            \node[latent, right=of g] (m) {$\boldsymbol\mu$}; %
            \node[latent, above=of g] (z) {$z$}; %
            \node[latent, above=of m] (p) {$\boldsymbol\Lambda$}; %
            % Plate
            \plate {} {(g)(z)} {$N$}; %
            % Edges
            \edge {m} {g};
            \edge {p} {g, m};
            \edge {z} {g};
        \end{tikzpicture}
        \caption{Likelihood Family}
        \label{fig:likelihood-family}
    \end{subfigure}
    \hspace{.5em}
    \begin{subfigure}{.5\linewidth}
        \centering
        \begin{tikzpicture}
            % Nodes
            \node[obs] (g) {$g$}; %
            \node[obs, left=of g, xshift=0.4cm] (s) {$s$}; %
            \node[obs, left=of s, xshift=0.4cm] (x) {$\mathbf{x}_{\leq H}$}; %
            \node[latent, above=of s, yshift=0.5cm] (m) {$\boldsymbol\mu$}; %
            \node[latent, above=of g, yshift=-0.5cm] (z) {$z$}; %
            \node[latent, above=of x, yshift=-0.5cm] (p) {$\boldsymbol\Lambda$}; %
            % Plate
            \plate {} {(g)(x)(s)(p)(m)(z)} {$N$};
            % Edges
            \edge {x} {p, m};
            \edge {x} {p};
            \edge {s} {m};
            \edge {m} {z};
            \edge {g} {z};
            \edge {p} {m};
            \edge[bend right=80] {p} {z};
        \end{tikzpicture}
        \caption{Variational Family}
        \label{fig:posterior-family}
    \end{subfigure}
    \caption{Graphical model for the GNeVA showing the likelihood family (left) and variational family (right).}
    \label{fig:spatial-distribution}
\end{figure}

\begin{proposition}
    \label{prop:multi-modality}
    Goals are multi-modal samples from a mixture of diverse intention distributions, and a single observed goal in the data is a sample from one dominant intention at a specific timestamp.      
\end{proposition}
Following this proposition, we leverage a mixture indicator variable $z$ to learn which intention mixture component has the dominant effect on the goal. In practice, $z_n\in\left\{0, 1\right\}^C$ is a one-hot vector sampled from a Categorical distribution parameterized by $\pi$. 
\begin{proposition}
    \label{prop:uncertainty-attribution}
    The distribution of goals reflects a joint effect of certainty and uncertainty, where certainty is associated with expectations, and the covariances quantify uncertainty.
\end{proposition}
Following this proposition, the mixture components are bivariate Gaussian distributions parameterized by expectations $\boldsymbol{\mu}$ and precisions $\boldsymbol{\Lambda}$ (i.e., \textit{inverse covariances}). In previous works, parameterization is accomplished by either training a neural network to output the point estimations of the mixture distribution parameters~\citep{pmlr-v100-chai20a} or using ensemble methods to approximate the covariances~\citep{9812107}. The common issue is that the output point estimations from the neural networks are optimized using training data. For the unseen or OOD cases, the same set of parameters may no longer be effective, hinging its capability to generalize across scenarios.

We address this limitation by adopting a Bayesian mixture with conjugate priors to account for the epistemic uncertainty of the parameters. The conjugate priors can bring algebraic convenience and closed-form expression to the posterior. We consider the means for all mixture components $\boldsymbol{\mu}$ to follow a bivariate normal prior, where prior mean $\eta_0$ and parameter $\beta_0$ are trainable. The precisions $\boldsymbol{\Lambda}$ are said to follow a Wishart distribution with trainable parameters $V_0$ and $\nu_0$. 

However, directly learning parameters of the true posterior $p(\boldsymbol{\mu,\Lambda\mid g,z})$ from the training dataset can not ensure the model's generalizability. The learned parameters can only express the belief of the latent parameters under the training dataset. Instead, we propose to parameterize a variational posterior distribution $q(\boldsymbol{\mu, \Lambda\mid g,z},s,\mathbf{x}_{\leq H})$ leveraging a \textit{causal structure} between environment and goal.

\paragraph{Mean Posterior} Since the expectations should reflect certainty, the variational mean posterior distribution reflects the belief of the expected goal location conditioned by observation. In real-world cases, the posterior variable $\eta_c$ is strongly bound to the concept of "anchors." For example, if we observed the lane where the target vehicle was positioned was a left-turn lane (contained in feature $\mathbf{s}$). The downstream left-turn exit lane was empty (contained in feature $\mathbf{x}_{\leq H}$), the most probable goal location somewhere on the exit lane. Meanwhile, $\boldsymbol\beta_c$ measures our belief in the uncertainty about this distribution, which the observations of the environment and interaction history should also deduce.

 \paragraph{Precision Posterior} Different from the mean posterior distribution, the precision posterior is conditioned only on the history traffic states $\mathbf{x}_{\leq H}$. The reason is two-folded:
\begin{itemize}
    \item We argue that uncertainty is rooted in the ever-changing dynamics of the surroundings rather than the static components. Since the environment is considered static in our problem settings, we drop them from the condition to avoid spurious correlations.
    \item Meanwhile, uncertainty quantification requires anticipation of the plausible future. To support the anticipation, we incorporate the full history feature as the function input.
\end{itemize}

Therefore, the joint variational posterior of the latent random variables is in the form
\begin{equation}
    \label{eq: variational}
    \prod\limits_{n=1}^N\prod\limits_{c=1}^C\mathcal{N}(\boldsymbol\mu_c\mid\boldsymbol\Lambda_c,\mathbf{x}_{\leq H},\mathbf{s})\mathcal{W}(\boldsymbol{\Lambda}_c\mid\mathbf{s})q(z_{nc}).
\end{equation}
where the $z$-posterior is approximated by
\begin{equation}
    \label{eq: z-posterior}
    q(z_{nc})=\frac{\pi_i\mathbb{E}_{q(\boldsymbol\mu_c,\boldsymbol\Lambda_c)}\left[\log p(g_n\mid\boldsymbol\mu_c,\boldsymbol\Lambda_c)\right]}{\sum\limits_{i=1}^C\pi_i\mathbb{E}_{q(\boldsymbol\mu_i,\boldsymbol\Lambda_i)}\left[\log p(g_n\mid\boldsymbol\mu_i,\boldsymbol\Lambda_i)\right]}.
\end{equation}

Derivation of equation~\ref{eq: variational} and~\ref{eq: z-posterior} are in the supplementary material. By constructing our model this way, we decouple the goal observation with posterior means and precisions (see~Figure \ref{fig:posterior-family}), which benefits the performance in three ways. First, it prevents the neural network for estimating posterior parameters from overfitting the distribution of goals in the training dataset, which helps guarantee the generalization performance. Meanwhile, since means and precisions are independent of the locations, we can train a proxy $z$-posterior network (see Section~\ref{sec: z-proxy}) to evaluate mixture assignment in unseen scenarios. Finally, we can derive a closed-form objective for training our model (see Section~\ref{sec: training}).

%% file: sections/method/feature_encoder.tex
\subsection{Feature Encoding and Attention Module}
\label{sec: feature-encoder}

\paragraph{Feature Encoding} Following the approaches in existing works~\citep{gao2020vectornet, pmlr-v155-zhao21b}, we represent the observation of a traffic scenario as a collection of polylines consisting of map-related polylines and agent history trajectories. Each polyline is broken into a collection of vectors containing origin and destination coordinates, other attributes such as heading, velocity, and dimensions for trajectories, and polyline types for map-related polylines. We encode map polylines and agents' history trajectories using two separate polyline encoders, resulting in three features: map features $\mathbf{m}$, target traffic participant's history feature $\mathbf{e}$, and surrounding participants' history features, $\mathbf{o}$.

\paragraph{Attention Module} To capture global interaction and avoid spurious correlations, we use two separate attention modules: the \textit{context attention} and the \textit{interaction attention} module to \textit{map-target} and \textit{surrounding-target} interactions, using target history feature $\mathbf{e}$ as query, and $\mathbf{m}$ and $\mathbf{o}$ as key and value, respectively. Both modules contain a stack of multi-head attention encoders~\citep{NIPS2017_3f5ee243} followed by MLPs. The context feature is the sum of the output from the context and interaction attention module to jointly represent $(\mathbf{x}_{\leq H},\mathbf{s})$, whereas the interaction feature is the output from the interaction attention module. Details about the attention modules are listed in the supplementary.
\begin{table*}[!th]
    \centering
    \caption{Performance Results on Argoverse validation set.}
    \begin{tabular}{@{}lccc@{}}
        \toprule
        \multicolumn{1}{c}{} & \textbf{mADE$_6$} & \textbf{mFDE$_6$}        & \textbf{MR$_6$}            \\ \midrule
        TPCN~\citep{Ye_2021_CVPR}                 & \underline{0.73}   & 1.15          & 0.11          \\
        mmTrans~\citep{9812060}              & \textbf{0.71}   & 1.15          & 0.11          \\
        LaneGCN~\citep{10.1007/978-3-030-58536-5_32}             & \textbf{0.71}   & \underline{1.08}          & -             \\
        \textbf{GNeVA (Ours)}        & 0.78   & \textbf{1.06} & \textbf{0.10} \\ \bottomrule
    \end{tabular}
    \label{tab: argo-performance}
\end{table*}

%% file: sections/method/z_proxy.tex
\subsection{Proxy $z$-posterior Network}
\label{sec: z-proxy}
The Proxy $z$-posterior Network (z-proxy) aims to parameterize the mixture assignment over the components and serve as a proxy for the learned $z$-posterior $q(\boldsymbol{z}\mid \boldsymbol{g,\mu,\Lambda})$. We model the $z$-proxy with an MLP conditioned on the context feature $\tilde{p}(\boldsymbol{z}|\mathbf{x}_{\leq H},\mathbf{s})$. To be noticed, $z$-posterior and $z$-proxy are mathematically different, where the former associates a location in space with a mixture, while the latter estimates the assignment conditioned on the current environment and vehicle movement. Similar implementations exist in previous literature~\citep{Dendorfer_2021_ICCV} and have been shown to help maintain performance in unseen and OOD cases.

%% file: sections/method/sampling.tex
\subsection{Sampling and Trajectory Completion}
\label{sec: sampling}
Recall that each mixture component of the spatial distribution is a multivariate Gaussian with Normal-inverse-Wishart conjugate prior. The posterior predictive distribution for a newly observed location $g^\prime$ is in the form
\begin{equation}
    \label{eq: posterior-predictive}
        p(g^*)\approx\sum\limits_{c=1}^C\tilde{p}(\boldsymbol{z}|\mathbf{x}^\prime_{\leq H},\mathbf{s}^\prime)\tau_{\nu_c-1}\left(\eta_c, \frac{\beta_c + 1}{\beta_c(\nu_c-1)}V_c^{-1}\right),
\end{equation}
where $\tau(\cdot)$ is the multivariate student-t distribution, $\eta_c$ and $\beta_c$ are the mean and scaling factor of the mean posterior, $V_c$ and $\nu_c$ are the scale matrix and degree of freedom of the precision posterior. We weigh each component probability using output from the $z$-proxy network. The goals are sampled using the Non-maximum Suppression (NMS), as described in Algorithm ~\ref{alg:nms}. After we obtain the list of goal candidates, we use a trajectory network modeled by a cascade of MLPs to predict and complete the intermediate trajectory given the goal candidates and the context feature.

%% file: sections/method/training.tex
\subsection{Model Training}
\label{sec: training}
We train our the generative model by maximizing the log-evidence lower bound (ELBO), given by
\begin{equation}
    \label{eq: ELBO loss}
    \begin{aligned}
         \mathcal{L}_\text{ELBO} &= \sum\limits_{c=1}^Cq(\boldsymbol{z})\mathbb{E}_{q(\boldsymbol{\mu},\boldsymbol{\Lambda})}\left[\log p(g|\boldsymbol{\mu},\boldsymbol{\Lambda},z)\right] \\
         &- D_{KL}\left[q(\boldsymbol{\mu,\Lambda})|p(\boldsymbol{\mu,\Lambda})\right] \\
         &- \mathbb{E}_{q(\boldsymbol{\Lambda})q(\boldsymbol{\mu})}D_{KL}\left(q(z|\boldsymbol{\mu},\boldsymbol{\Lambda})|p(\boldsymbol{z})\right),
    \end{aligned}
\end{equation}
where $D_{KL}(\cdot)$ is the Kullback–Leibler divergence. Please refer to the supplementary material for the closed-form ELBO loss. Meanwhile, we train the z-proxy network by minimizing the cross-entropy
\begin{equation}
    \label{eq: Cross-Entropy}
    \mathcal{L}_{z} = -\mathbb{E}_{q(\boldsymbol{z})}\left[\log\tilde{p}(\boldsymbol{z})\right].
\end{equation}
The trajectory network is trained to minimize the Huber Loss between the predicted trajectory and ground-truth trajectory given the ground-truth goal. In practice, we train the spatial distribution model and the trajectory network separately.

%% file: sections/experiment.tex
\begin{figure*}[!th]
    \centering
    \begin{subfigure}{0.45\linewidth}
        \caption{In-distribution case: \texttt{CHN\_Merging\_ZS0}}
        \includegraphics[width=\textwidth]{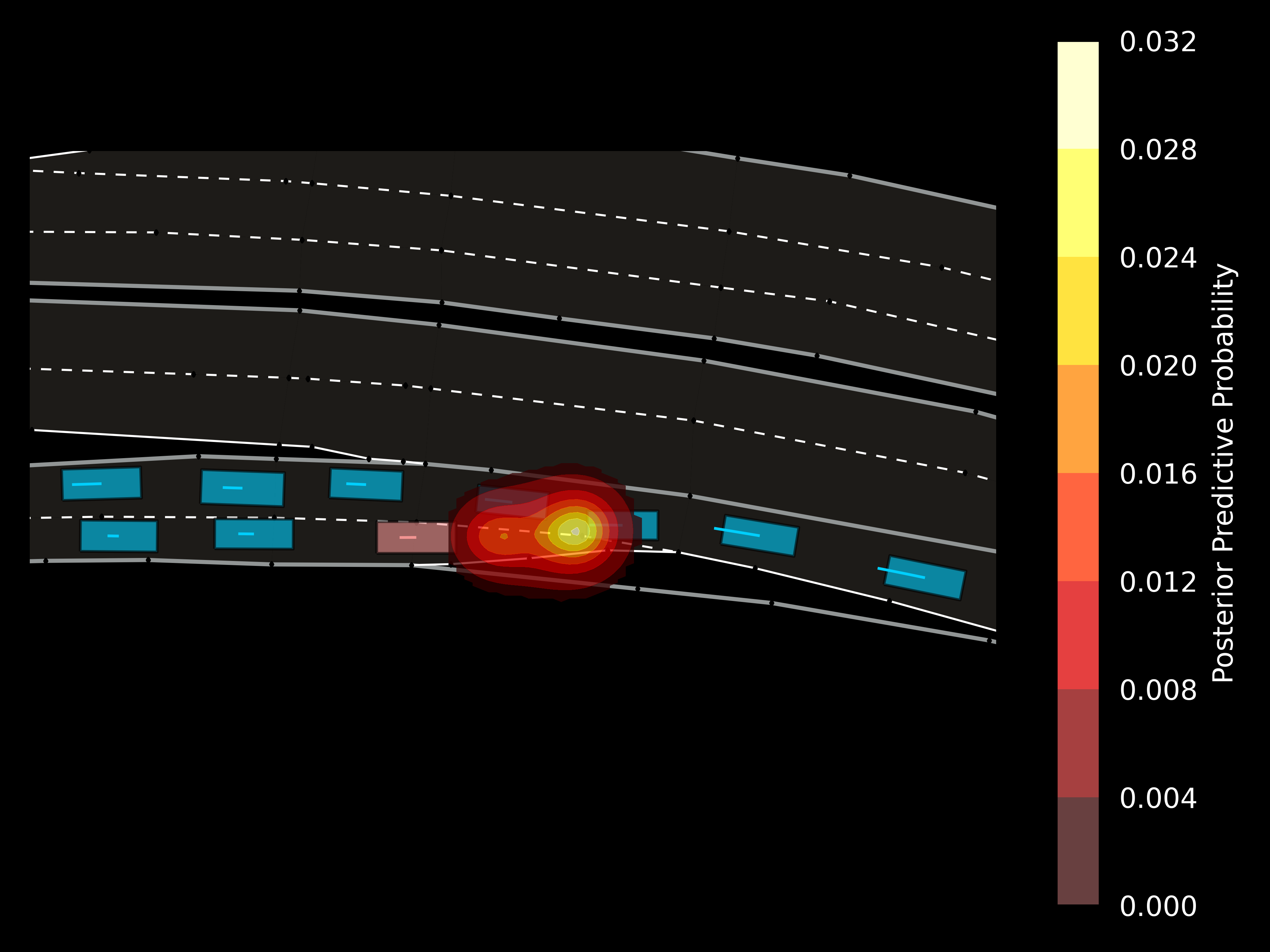}
        \label{fig: merging-id}
    \end{subfigure}
    \hspace{0.1em}
    \begin{subfigure}{0.45\linewidth}
        \caption{OOD case: \texttt{Merging\_TR0}}
        \includegraphics[width=\textwidth]{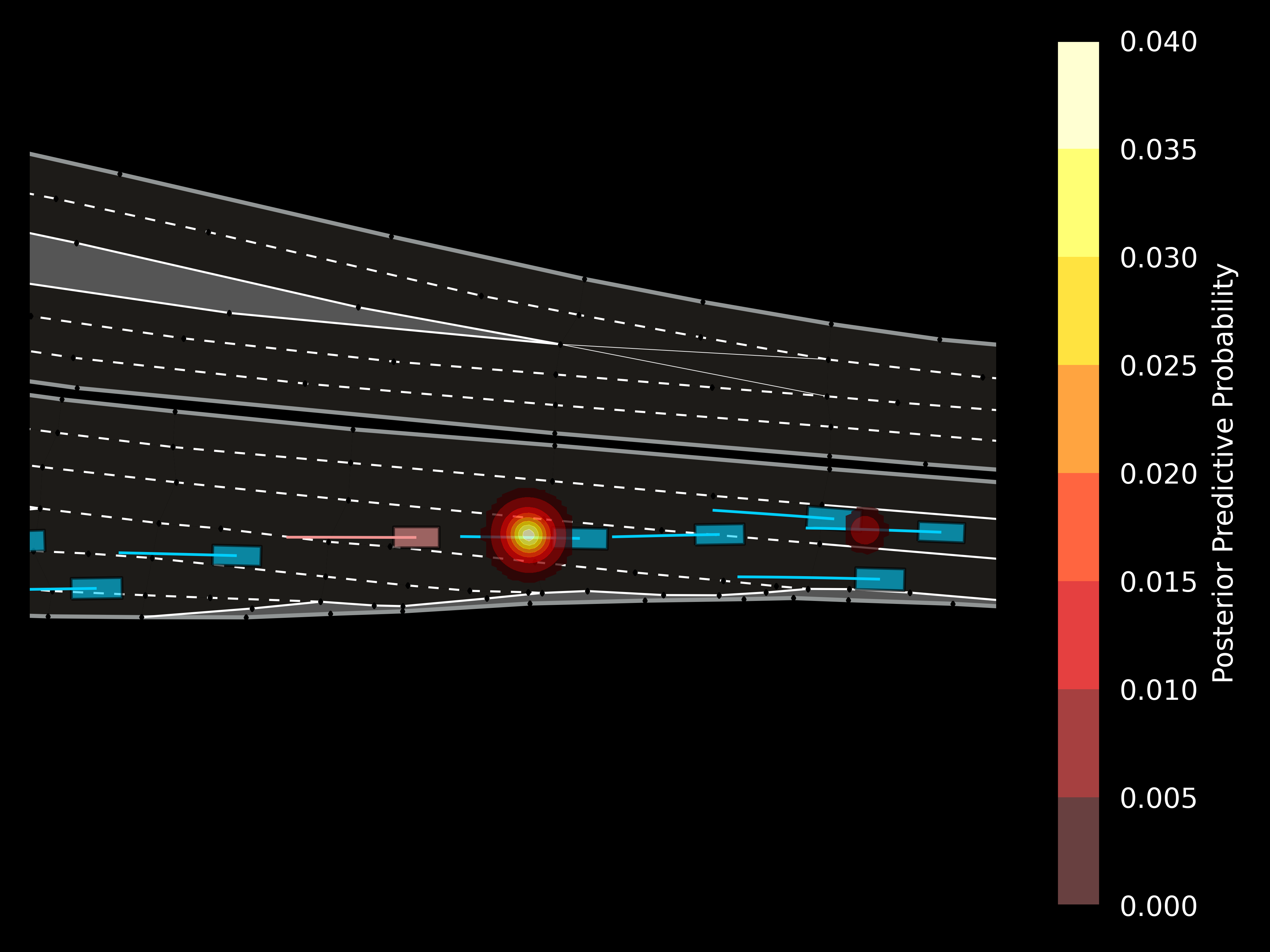}
        \label{fig: merging-ood}
    \end{subfigure}
    \label{fig: merging}
    \vspace{-0.5em}
    \begin{subfigure}{0.45\linewidth}
        \caption{In-distribution case: \texttt{USA\_Intersection\_MA}}
        \includegraphics[width=\textwidth]{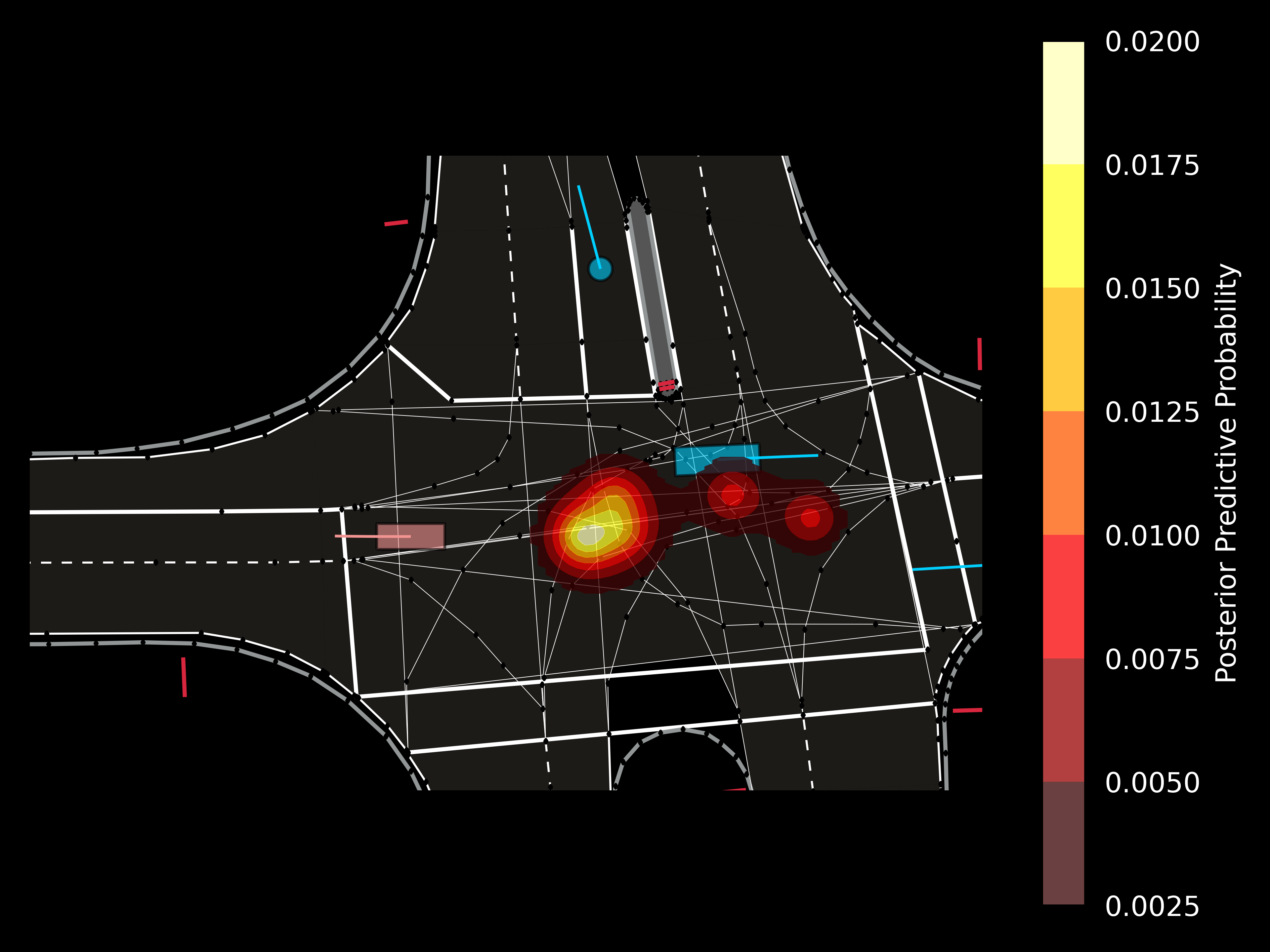}
        \label{fig: intersection-id}
    \end{subfigure}
    \hspace{0.1em}
    \begin{subfigure}{0.45\linewidth}
        \caption{OOD case: \texttt{Intersection\_CM}}
        \includegraphics[width=\textwidth]{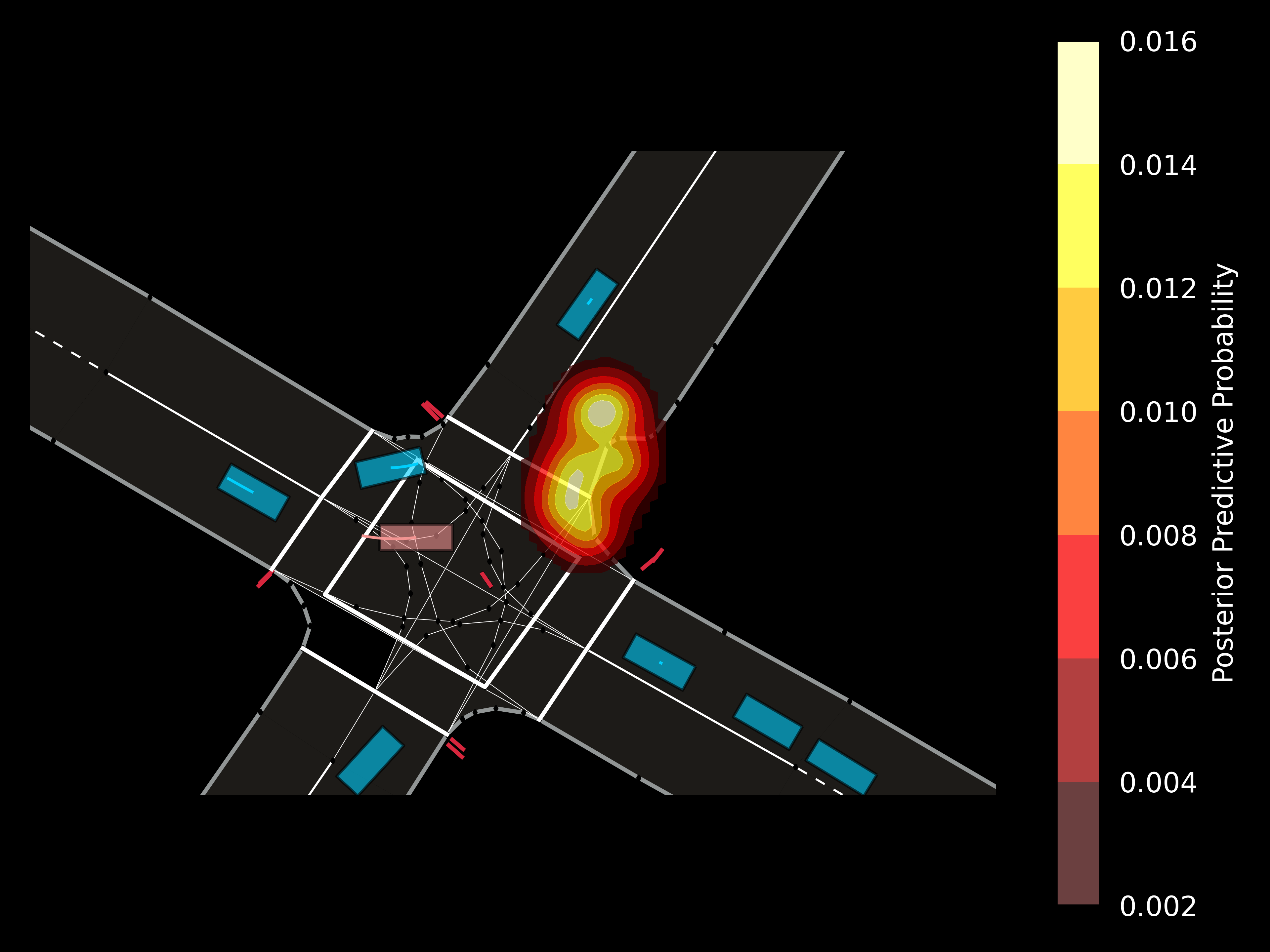}
        \label{fig: intersection-ood}
    \end{subfigure}
    \vspace{-0.5em}
    \begin{subfigure}{0.45\linewidth}
        \caption{In-distribution case: \texttt{USA\_Roundabout\_SR}}
        \includegraphics[width=\textwidth]{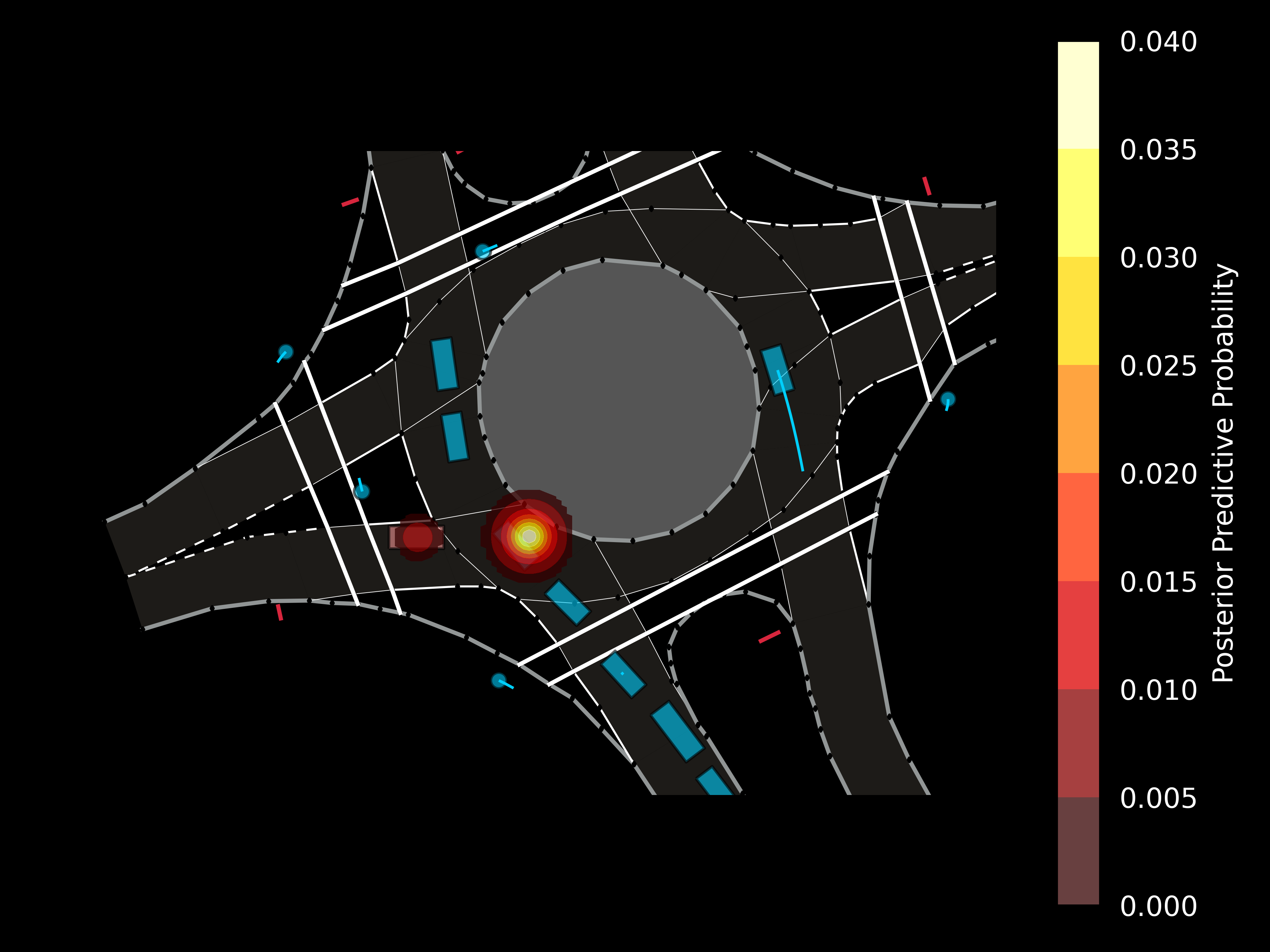}
        \label{fig: roundabout-id}
    \end{subfigure}
    \hspace{0.1em}
    \begin{subfigure}{0.45\linewidth}
        \caption{OOD case: \texttt{Roundabout\_RW}}
        \includegraphics[width=\textwidth]{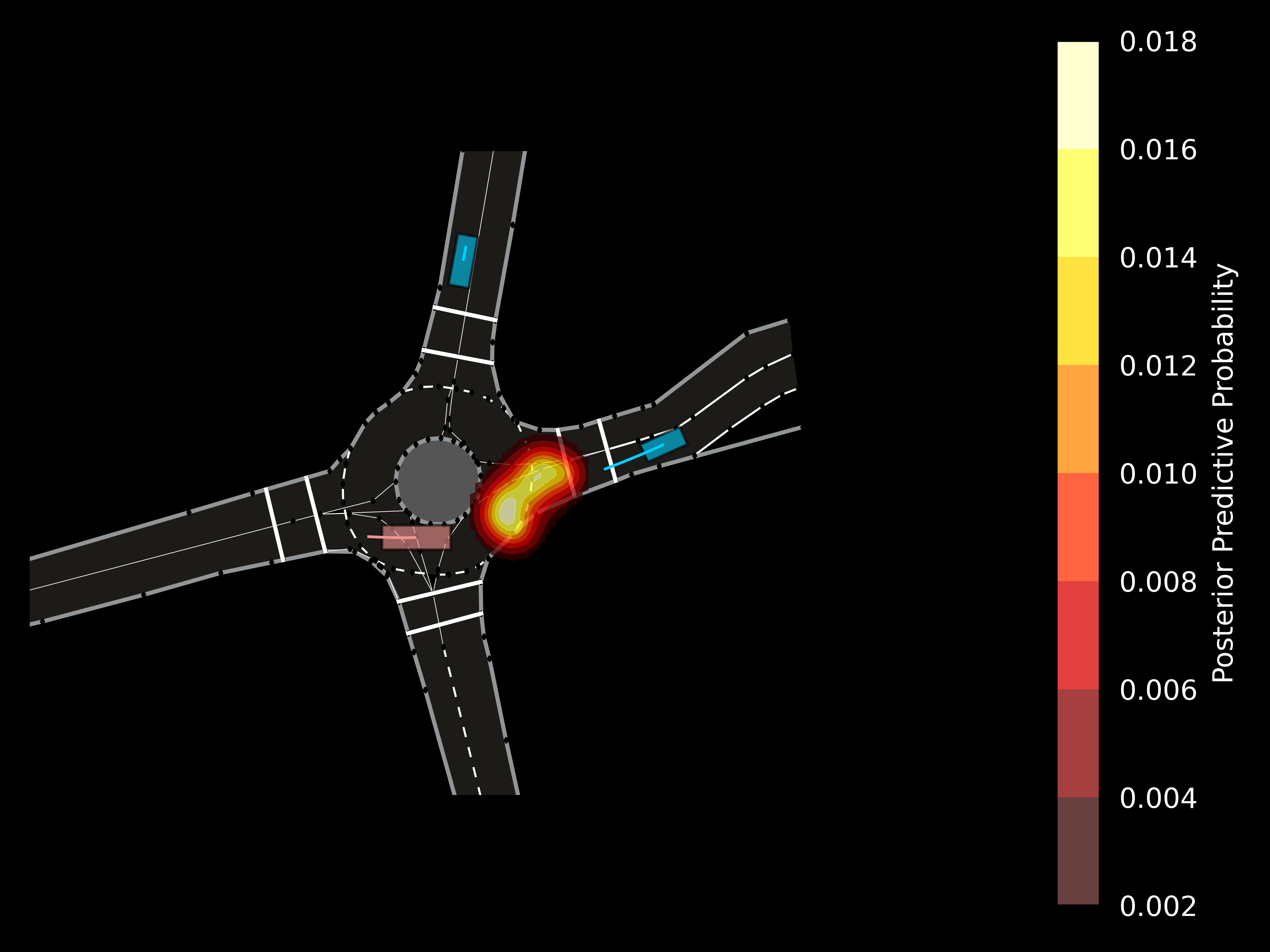}
        \label{fig: roundabout-ood}
    \end{subfigure}
    \caption{Visualization of posterior predictive goal distributions under selected in-distribution and out-of-distribution cases. All cases are selected from the INTERACTION test dataset.}
    \label{fig: qualitative}
\end{figure*}%

% =================================================================================================
We empirically evaluate the GNeVA on the popular motion prediction datasets following the experiment settings introduced in section~\ref{sec: experiments}. Section~\ref{sec: benchmark} presents performance comparison and analysis with state-of-the-art motion prediction methods. In section~\ref{sec: generalization}, we evaluate the model's generalization ability. Finally, section~\ref{sec: qualitative} presents visualizations of predictive distribution in in-distribution and out-of-distribution cases with qualitative analysis of the model's interpretability.

\subsection{Experiment Settings}
\label{sec: experiments}

\subsubsection{Dataset}
\label{sec: datasets}
We evaluate our model in single-agent motion prediction settings on the Argoverse Motion Forecasting dataset~\citep{Argoverse} and the INTERACTION dataset~\citep{https://doi.org/10.48550/arxiv.1910.03088}. The Argoverse consists of roughly $323,557$ scenarios, the task of which is to predict future three-second motions given the HD map and the two-second history of the agent. The INTERACTION dataset consists of around $398,409$ cases, and the objective is to predict the future trajectory of target vehicles in three consecutive three seconds given one-second history observations. Additionally, the INTERACTION dataset provides manual labels of different road geometry, including highway merging, intersections, and roundabouts, which is convenient for evaluating OOD performance.

\subsubsection{Metrics}
\label{sec: metrics}
We use standard motion prediction metrics, including the minimum average displacement error (mADE$_k$), minimum final displacement error (mFDE$_k$), and the miss rate (MR$_k$). The mADE$_k$ and mFDE$_k$ are calculated between top-$k$ trajectory predictions and the ground truth, given as
\begin{equation}
    \text{mADE}_k = \min\limits_k\frac{1}{T}\sum\limits_{t=H+1}^{H+T}\sqrt{(\hat{x}_t^k-x_t)^2 + (\hat{y}_t^k-y_t)^2},
\end{equation}
\begin{equation}
    \text{mFDE}_k = \min\limits_k\sqrt{(\hat{x}_{H+T}^k - x_{H+T}^k) + (\hat{y}_{H+T}^k - y_{H+T}^k)},
\end{equation}%
where $T$ is the number of time steps needed to predict the future. MR$_k$ is only calculated following the definition given by the Argoverse dataset. It is the ratio of cases where minFDE$_k$ is higher than 2 meters.
% For the INTERACTION dataset, we first define a lateral and a longitudinal displacement threshold. The lateral threshold is set as 2 meters. The longitudinal displacement threshold is a piece-wise linear function about the last observed ground-truth velocity $v_H$ of the vehicle to predict, given as,
% \begin{equation}
%     \text{Threshold}_\text{lng} = \begin{cases}
%         1, & v_H < 1.4\text{m/s}^2 \\
%         1 + \frac{v_H - 1.4}{9.6}, & 1.4\text{m/s}^2\leq v_H<11\text{m/s}^2 \\
%         2, & v_H\geq 11\text{m/s}^2
%     \end{cases}.
% \end{equation}
% A case of prediction is considered a "miss" if all of its top-$k$ trajectory prediction endpoints land outside of the bounding box centered at the ground-truth endpoint with dimensions given by the longitudinal and lateral thresholds. The MR$_k$ is the ratio of "miss" cases in all test cases.

\subsection{Benchmark Results}
\label{sec: benchmark}

\begin{table}[!ht]
    \centering
    \caption{Results on INTERACTION validation set.}
    \begin{tabular}{@{}lcc@{}}
    \toprule
                            & \textbf{mADE$_6$} & \textbf{mFDE$_6$}  \\ \midrule
        DESIRE~\citep{Lee_2017_CVPR}              & 0.32       & 0.88    \\
        MultiPath~\citep{pmlr-v100-chai20a}           & 0.30       & 0.99    \\
        TNT~\citep{pmlr-v155-zhao21b}                 & \textbf{0.21}       & \underline{0.67}    \\
        \textbf{GNeVA (Ours)}       & \underline{0.25}       & \textbf{0.64} \\
        \bottomrule
    \end{tabular}
    \label{tab: interaction-performance}
\end{table}

We compare the proposed GNeVA motion prediction model with the state-of-the-art method on the validation split of both datasets. As shown in Table~\ref{tab: argo-performance}, our model achieves state-of-the-art performance measured by minFDE$_6$ and miss rate on the Argoverse dataset, with a slightly higher (i.e., $\approx 0.08$ meters) minADE$_6$ compared to the others. Table~\ref{tab: interaction-performance} shows the performance result evaluated on the INTERACTION validation set. Our model has the best mFDE$_6$ and the second-place mADE$_6$ performance among the four models compared. The GNeVA model can achieve performance comparable to state-of-the-art motion prediction models.

\begin{table*}[!th]
\centering
\caption{Model Performance under Cross-scenario Tests}
\label{tab: gen-test}
\begin{tabular}{@{}ccccccc@{}}
\toprule
                      & \multicolumn{6}{c}{\textbf{Train Scenario}}                                                                                             \\ \midrule
                      & \multicolumn{2}{c}{\textbf{Intersection}}   & \multicolumn{2}{c}{\textbf{Roundabout}}     & \multicolumn{2}{c}{\textbf{Full Dataset}}           \\
\textbf{Validate Scenario}      & \textbf{mADE$_6$} & \textbf{mFDE$_6$} &  \textbf{mADE$_6$} & \textbf{mFDE$_6$} &  \textbf{mADE$_6$} & \textbf{mFDE$_6$}  \\ \midrule
\textbf{Intersection} & 0.56          & 1.41          & 0.56          & 1.39          & 0.31          & 0.73          \\
\textbf{Roundabout}   & 0.61          & 1.56          & 0.44          & 1.08          & 0.32          & 0.76          \\ \bottomrule
\end{tabular}
\end{table*}
\begin{table*}[!th]
    \centering
    \caption{Cross Dataset Evaluation Results.}
    \label{tab: cross-dataset}
    \begin{tabular}{@{}lccccc@{}}
        \toprule
        \multicolumn{1}{c}{\textbf{Dataset}} & \multicolumn{3}{c}{\textbf{Argoverse (validate)}} & \multicolumn{2}{c}{\textbf{INTERACTION (validate)}} \\ \cmidrule(l){2-6} 
        \multicolumn{1}{c}{} & \textbf{mADE$_6$}  & \textbf{mFDE$_6$} & \textbf{MR$_6$}   & \textbf{mADE$_6$}   & \textbf{mFDE$_6$}   \\ \midrule
        \textbf{Argoverse (train)}    & 0.78        & 1.06       & 0.10     & 0.37         & 0.91         \\
        \textbf{INTERACTION (train)}  & 0.92        & 1.34       & 0.15     & 0.25         & 0.64        \\ \bottomrule
    \end{tabular}
\end{table*}
\subsection{Generalization Evaluation}
\label{sec: generalization}

\subsubsection{Cross-scenario Generalizability}
The model's generalization ability can be evaluated directly through a cross-scenario test~\citep{9981096}, that is, to mimic the distribution shift in a real scenario by training and testing the model under different traffic conditions. The drop in performance can reflect the existence of generalization bottlenecks. Table~\ref{tab: gen-test} shows the cross-scenario test results. From the results, we find that the model that is trained solely using data collected from the intersections can achieve similar performance under out-of-domain cases (roundabout), with about 8\% in mADE$_6$ and 10\% in mFDE$_6$. Meanwhile, the model trained solely using data collected from the roundabout can also maintain its performance in OOD cases, with only a 5\% % increase in MR$_6$ overall. However, since the total number of interaction cases and roundabout cases is relatively small, with the same amount of training steps, models trained on these subsets can't achieve the same performance compared to those trained on the full dataset. In summary, despite the significantly reduced number of training data and differences between road geometry, the GNeVA model can maintain its performance when directly applied to an unseen scenario, showcasing promising cross-scenario generalizability.

\subsubsection{Cross-dataset Generalizaility}

Evaluating the generalization performance under different datasets is more challenging than cross-scenario evaluation since distribution shifts exist in underlying traffic conditions and noise due to different sensors. The evaluation results are listed in Table~\ref{tab: cross-dataset}. We observed a $0.14$ meters increase in mADE$_6$, a $0.28$ meters increase in mFDE$_6$, and a $5\%$ increase in MR$_6$ when trained on INTERACTION and evaluated on the Argoverse. On the other hand, the mADE$_6$ and mFDE$_6$ are still below $0.5$ meters and $1.0$ meters when trained on the Argoverse and evaluated on the INTERACTION. Therefore, the performance degradation when the model is trained on the INTERACTION dataset and evaluated on the Argoverse dataset is higher than the opposite. This fits our expectation that the Argoverse dataset has a higher diversity and is collected under a wider urban area than the INTERACTION dataset. Results from this experiment show that our model has the potential generalizability to apply to unseen datasets directly but can face challenges when trained on a significantly less diverse training set. We argue that such a limitation can potentially be rooted in the fact that the neural network used to parameterize the posterior fails to learn a sufficiently generalizable mapping from observed data to posterior parameters.

\subsection{Qualitative Analysis}
\label{sec: qualitative}

Since the GNeVA model aims to improve interpretability by directly modeling the distribution of goals, it is essential to validate if the predicted distribution fits the properties of driver intention. In Figure~\ref{fig: qualitative}, we select and visualize six cases from the INTERACTION test dataset, where three of them are in-distribution cases with seen road geometry and similar driver behavior in the training dataset, whereas the other three cases are collected from locations unseen in the training dataset.

In the first row, we compare the cases where the target vehicle is merging into the mainline of a highway section. Plots indicate that our model can successfully distinguish the lanes in the ramp and the mainline and identify the staying-within-ramp intention in the in-distribution case (see Figure~\ref{fig: merging-id}), as well as the merging-left-into-mainline intention in the out-of-distribution case (see Figure~\ref{fig: merging-ood}). More importantly, we observe two separate distribution modes in two adjacent lanes. This indicates that the model successfully anticipates two lane-changing intentions: finish the current lane-change maneuver, stay in the lane, or accelerate and change to an inner adjacent lane for higher speed.

The middle row shows results on two different intersections. We compare two similar cases with potential left-turn target vehicles but at different locations and with different current headings. In the in-distribution case (see Figure~\ref{fig: intersection-id}), the distribution has three clear clusters: the right-most one is associated with going straight, the middle mode potentially allows the target vehicle to turn left and enter the outer lane of the downstream exit, and the left-most cluster shows a diverse intention, including going straight or turning left and entering the inner lane of the exit. In the OOD case (see Figure~\ref{fig: intersection-ood}), the multi-modal distribution concentrates along the only left-turn exit lane since the target vehicle has a clear left-turn heading. The two examples showcase the power of GNeVA to identify multiple plausible intentions, which address the multi-modal property of driver intention.

Visualizations of the two cases in the bottom row demonstrate predictive distributions on two different roundabouts. As shown in Figure~\ref{fig: roundabout-id}, the model can successfully anticipate two plausible intentions: staying behind the stop line and yield (i.e., \textit{the left mode}) and accelerating to enter the roundabout (i.e., \textit{the right mode}). In Figure~\ref{fig: roundabout-ood}, we choose the OOD case where there is a small roundabout significantly different from the common large-scale roundabouts in the training data. Despite the model predicting a predictive distribution that fits the constraints of the roundabout geometry, we see that it only identifies the plausible path toward the nearest exit, which shows that the model can be further improved in the future to tackle cases with road geometry that is rare in the training data. Further experiments evaluating the model's sensitivity to road geometry can be found in the supplementary. 

% \subsection{Qualitative Analysis}
% \label{sec: qualitative}

% \subsection{Ablation Study}
% \label{sec: ablation}

% \paragraph{Sampling radius} Choosing a proper sampling radius can prevent under- or over-estimate the importance of 
% \begin{table}[htp]
% \centering
% \caption{Comparison of different NMS sampling radius}
% \label{tab: sampling-radius}
% \begin{tabular}{@{}ccc@{}}
% \toprule
%  Sampling radius (m) & mFDE & MR \\ \midrule
%  &      &    \\
%  &      &    \\
%  &      &    \\ \bottomrule
% \end{tabular}
% \end{table}

% \paragraph{Sampling IoU Threshold} We investigate how different IoU threshold settings can affect the prediction performance. As shown in Table~\ref{tab: iou-threshold}, 
% \begin{table}[htp]
% \centering
% \caption{Comparison of different NMS IoU thresholds}
% \label{tab: iou-threshold}
% \begin{tabular}{@{}ccc@{}}
% \toprule
% IoU Thresholds (\%) & mFDE & MR \\ \midrule
%  &      &    \\
%  &      &    \\
%  &      &    \\ \bottomrule
% \end{tabular}
% \end{table}

%% file: sections/conclusion.tex
In this paper, we propose the Goal-based Neural Variational Agent (GNeVA), a deep variational Bayes model that evaluates the spatial distribution of the long-term goals for drivers. We postulate the propositions about the multi-modality and a causal structure associated with the goal. Following these propositions, we design the model using a variational mixture of Gaussian distributions with posterior parameterized by the neural network. In the experiment, our model achieved comparative performance as the state-of-the-art models, showcased a promising generalization ability to the unseen cases, and demonstrated the ability to generate predictive distributions that address the constraints from road geometry and reflect the multi-modal property of driver behavior.

\paragraph{Limitations} The GNeVA model is inherently a single-agent trajectory prediction model and requires global coordinates to be first projected into the local coordinate frame, which can lead to redundant data preprocessing for multi-agent prediction. Therefore, how to efficiently scale the model to predict multiple agents simultaneously can be an exciting topic. Besides, we reduce the problem by building a goal distribution only conditioned on a static map and the history trajectories. How to handle dynamic environment states and incorporate a single agent's anticipations of the others' future motion as a variable in the model is worth exploring in the future.